\documentclass[10pt]{article} % For LaTeX2e
\usepackage[preprint]{tmlr}

\usepackage{amsmath,amsfonts,bm}

\def\eqref#1{equation~\ref{#1}}
\def\1{\bm{1}}

\DeclareMathAlphabet{\mathsfit}{\encodingdefault}{\sfdefault}{m}{sl}
\SetMathAlphabet{\mathsfit}{bold}{\encodingdefault}{\sfdefault}{bx}{n}

\usepackage{hyperref}
\usepackage{url}

\usepackage{booktabs}       % professional-quality tables
\usepackage{amsfonts}       % blackboard math symbols
\usepackage{nicefrac}       % compact symbols for 1/2, etc.
\usepackage{microtype}      % microtypography
\usepackage{xcolor}         % colors
\usepackage{amsmath}
\usepackage{graphicx}
\usepackage[breakable]{tcolorbox}  % prompt boxes
\usepackage{CJKutf8}        % Chinese characaters in pdfLatex
\usepackage{amssymb} % for \checkmark
\usepackage{multirow}       % for multirows in tables
\usepackage{mathtools}
\definecolor{darkblue}{rgb}{0, 0, 0.5}
\hypersetup{colorlinks=true, citecolor=darkblue, linkcolor=darkblue, urlcolor=darkblue}

\DeclarePairedDelimiter{\norm}{\lVert}{\rVert}

\newcommand\methodname{\textcolor{black}{\textsc{Robust Rubric-Agnostic Reward}}}

\newcommand\shortmethodname{\textcolor{black}{\textsc{R3}}}

\title{$\shortmethodname$: Robust Rubric-Agnostic Reward Models}

\author{\name David Anugraha \email david.anugraha@stanford.edu \\
      \addr Department of Computer Science, Stanford University
      \AND 
      \name Zilu Tang \email zilutang@bu.edu \\
      \addr Department of Computer Science, Boston University
      \AND
      \name Lester James V. Miranda \email ljm@allenai.org \\
      \addr Allen Institute for AI 
      \AND 
      \name Hanyang Zhao \email hz2684@columbia.edu \\
      \addr Department of IEOR, Columbia University
      \AND
      \name Shou-Yi Hung \email rayh@cs.toronto.edu \\
      \addr University of Toronto
      \AND
      \name Mohammad Rifqi Farhansyah \email mrifqifarhansyah@gmail.com \\
      \addr Institut Teknologi Bandung
      \AND 
      \name Garry Kuwanto \email gkuwanto@bu.edu \\
      \addr Department of Computer Science, Boston University
      \AND
      \name Derry Wijaya \email wijaya@bu.edu \\
      \addr Department of Computer Science, Boston University \\
      \addr Monash University Indonesia
      \AND
      \name Genta Indra Winata \email genta.winata@capitalone.com\\
      \addr AI Foundations, Capital One}

\def\month{MM}  % Insert correct month for camera-ready version
\def\year{YYYY} % Insert correct year for camera-ready version
\def\openreview{\url{https://openreview.net/forum?id=XXXX}} % Insert correct link to OpenReview for camera-ready version

\begin{document}

\maketitle

\begin{abstract}
Reward models are essential for aligning language model outputs with human preferences, yet existing approaches often lack both controllability and interpretability. These models are typically optimized for narrow objectives, limiting their generalizability to broader downstream tasks. Moreover, their scalar outputs are difficult to interpret without contextual reasoning. To address these limitations, we introduce $\shortmethodname$, a novel reward modeling framework that is rubric-agnostic, generalizable across evaluation dimensions, and provides interpretable, reasoned score assignments. $\shortmethodname$ enables more transparent and flexible evaluation of language models, supporting robust alignment with diverse human values and use cases. Our models, data, and code are available as open source at~\url{https://github.com/rubricreward/r3}.
% at~\url{https://anonymized-link}.
\end{abstract}

\section{Introduction}
Reward models play a central role in aligning language model outputs with human preferences by assigning scalar scores to generated responses~\citep{rafailov2023direct, lambert2024rewardbench}. However, current reward modeling approaches suffer from two significant limitations: limited controllability and poor interpretability. First, these models are often optimized for narrow objectives—such as helpfulness or harmlessness—resulting in behavior that is overly tailored to specific metrics and not readily generalizable to a broader range of downstream tasks \citep{li2019acute, stureborg2024large}. Second, the interpretability of reward scores remains unclear. For instance, scalar values like “1” or “2” on a Likert scale are not inherently meaningful without an explicit explanation of what those scores represent in context.

Aligning models with human preferences is crucial, but obtaining human judgments is often costly and time-consuming~\citep{vu2024foundational,lin2025data,winata2025preference}. Leveraging existing human evaluations from prior research appears promising; however, it poses several challenges, including lack of standardization, varying evaluation criteria, insufficient documentation, data privacy issues, and proprietary restrictions~\citep{kim2024biggen}. As an alternative, using model-generated outputs for reward modeling or annotation offers greater efficiency and flexibility. This lack of generalizability and transparency presents challenges for reliably evaluating and guiding language model behavior across diverse use cases.

\begin{figure*}[!t]
    \centering
    \includegraphics[width=0.65\textwidth]{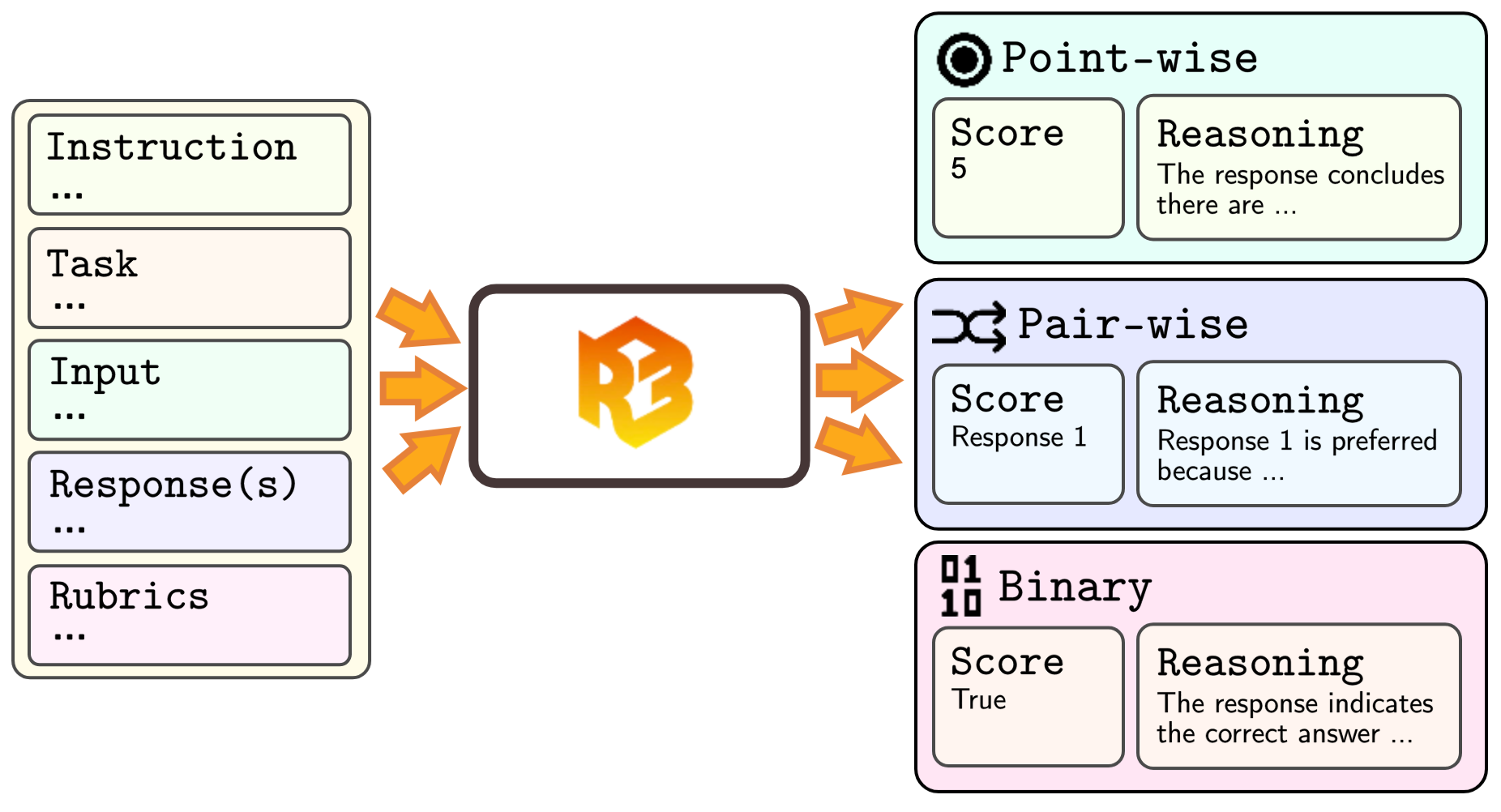}
    \caption{$\methodname{}$ ($\shortmethodname{}$) models both the input and output of a task. It takes a prompt that includes an instruction, task description, input, response(s), and evaluation rubrics, and generates a score along with the corresponding reasoning.}
    \label{fig:architecture}
    % \vspace{-2mm}
\end{figure*}

To address these issues, we propose $\shortmethodname{}$, a rubric-agnostic reward modeling framework that generalizes across evaluation dimensions while remaining grounded in interpretable, measurable scores. $\shortmethodname{}$ supports flexible alignment with human values and produces explicit reasoning for its score assignments, enabling more transparent and trustworthy evaluation. Here, ``rubric-agnostic'' refers to agnosticism toward the choice of criteria, not the absence of rubrics: unlike ArmoRM~\citep{wang2024interpretable}, which fixes a set of dimensions (e.g., helpfulness, harmlessness) at training time and requires retraining for new criteria, and unlike conventional scalar reward models, which encode criteria implicitly and opaquely in their parameters, $\shortmethodname{}$ takes an arbitrary natural-language rubric as explicit input at inference time. This makes the evaluation criteria transparent, inspectable, and customizable without retraining, allowing $\shortmethodname{}$ to evaluate responses against any provided rubric, not just the set of rubrics seen during training. Our contributions can be summarized as follows:
\begin{itemize}
    \item We introduce $\shortmethodname$, a novel task-agnostic robust reward model training framework that leverages fine-grained rubrics to provide highly controllable and interpretable reward scores. These rubrics can be either hand-crafted by humans or generated by LLMs.
    \item We propose a unified framework for training reward models by adapting various types of data into three standard formats: point-wise, pair-wise, and binary.
    \item We curate a new reward modeling $\shortmethodname{}$ dataset collected from 45 diverse sources that covers tasks such as classification, preference optimization, and question answering (Figure \ref{fig:datasets}). Each example in the dataset contains an instruction and task description, input, response(s), evaluation rubrics, and a score along with the corresponding reasoning (Figure \ref{fig:architecture}).
    \item We demonstrate that our $\shortmethodname$ models exhibit robust and superior performance, not only matching but often exceeding both established baselines and proprietary models across a diverse suite of tasks, including reward modeling, knowledge recall, reasoning, and summarization. Importantly, our framework maintains this high level of effectiveness even under stringent resource constraints—utilizing no more than 14,000 training examples and limited computational capacity—by leveraging efficient adaptation techniques such as low-rank adaptation (LoRA)~\citep{hu2022lora}.
\end{itemize}

\begin{table*}[!ht]
\centering
\caption{A comparison between existing models and $\shortmethodname{}$ across various dimensions, including data types, task formats, and evaluation rubrics. $^*$The model is neither closed-source nor proprietary.}
\resizebox{\textwidth}{!}{
    \begin{tabular}{l|cccc|ccc|c|c}
    \toprule
    \textbf{Method} & \multicolumn{4}{c|}{\textbf{Data}} & \multicolumn{3}{c|}{\textbf{Tasks}} & \textbf{Rubrics} & \textbf{Access}$^*$ \\
    & Size & Point-wise & Pair-wise & Binary & Point-wise & Pair-wise & Binary & Customizable \\
    \midrule
    ArmoRM~\citep{wang2024interpretable} & $\sim$974.4k & \checkmark & \checkmark & - & \checkmark & - & - & - & \checkmark \\
    CLoud~\citep{ankner2024critique} & $\sim$280k & \checkmark & - & - & \checkmark & - & - & - & \checkmark \\
    GenRM~\citep{zhang2024generative} & $\sim$157.2k & \checkmark & - & \checkmark & \checkmark & - & \checkmark & - & - \\
    JudgeLRM~\citep{chen2025judgelrm} & 100K &  \checkmark & - & - & \checkmark & \checkmark & - &  \checkmark &  \checkmark\\
    Prometheus1~\citep{kim2023prometheus} & 100k & \checkmark & - & - & \checkmark & \checkmark & - & \checkmark & \checkmark \\
    Prometheus2~\citep{kim2024prometheus2} & 300k & \checkmark & \checkmark & - & \checkmark & \checkmark & - & \checkmark & \checkmark \\ 
    m-Prometheus~\citep{pombal2025m} & 480k & \checkmark & \checkmark & - & \checkmark & \checkmark & - & \checkmark & \checkmark \\
    Self-Taught~\citep{wang2024self} & ? & - & 
    \checkmark & - & - & \checkmark & - & \checkmark & \checkmark \\
    SynRM~\citep{ye2024improving} & 5k & \checkmark & \checkmark & - & - & \checkmark & - & - & - \\
    UniEval~\citep{zhong2022towards} & $\sim$185.5k & - & - & \checkmark & - & - & \checkmark & \checkmark & \checkmark\\ 
    G-Eval~\citep{liu2023g} & ? & ? & ? & ? &\checkmark & \checkmark & \checkmark & \checkmark & - \\
    % MetaMetrics && & \checkmark & \checkmark & \checkmark & \checkmark \\ 
    FLAMe~\citep{vu2024foundational} & 5M+ & \checkmark & \checkmark & \checkmark & \checkmark & \checkmark & \checkmark & \checkmark & - \\
    RM-R1~\citep{chen2025rm} & $\sim$100k & - & \checkmark & - & - & \checkmark & - & \checkmark & \checkmark \\ \midrule
    % RM-R1 & - & \checkmark & - & \checkmark \\ \midrule
    \textbf{$\shortmethodname$} & \{4k, 14k\} & \checkmark & \checkmark & \checkmark & \checkmark  & \checkmark  &  \checkmark & \checkmark & \checkmark \\
    \bottomrule
    \end{tabular}
}
\label{tab:rm-comparison}
% \vspace{-2mm}
\end{table*}

\section{Aren't Existing Reward Models Robust Enough?}
\label{sec:2}

The challenge of building models that generalize across diverse tasks and domains—particularly in evaluating quality from multiple aspects or human annotation metrics—is well established. In this section, we present the motivation behind the need for developing new reward models.

\paragraph{Controllability.}
Existing reward models, such as ArmoRM~\citep{wang2024interpretable} and UniEval~\citep{zhong2022towards}, offer limited support for evaluating models on fine-grained aspects. They typically require separate training for each aspect along with corresponding parameter weights, reducing flexibility during both training and evaluation—especially when dealing with unseen aspects. Similarly, models like Prometheus~\citep{kim2023prometheus,kim2024prometheus2} are restricted in the range of supported task types; for example, they do not accommodate binary classification. ArmoRM is further limited in that it only supports point-wise tasks, making it unsuitable for pair-wise comparisons.

\paragraph{Interpretability.} Scores generated by reward models—particularly those based on generative LLMs \citep{skyworkcritic2024, yu2025improve} or custom classifiers \citep{wang2025helpsteerpreference, winata2024metametrics, zhang2024general} —can be difficult to interpret. For example, a score of 0.6543 on a 0–1 scale offers little clarity: Is it measuring helpfulness, correctness, coherence, or some opaque combination of all three? Without a clearly defined rubric or accompanying explanation, such scores provide limited actionable insight, leaving users to guess what aspect of quality the number is intended to capture.

\paragraph{Limited Compatibility on Various Tasks.}
Existing reward models often have limited compatibility with a diverse range of tasks. For instance, models like RM-R1~\citep{chen2025rm} are primarily designed for pair-wise comparisons, making them less suitable for point-wise or binary classification tasks, which limits their applicability. Similarly, Prometheus supports point-wise and pair-wise evaluations but lacks native support for binary classification—an approach that can be particularly effective for tasks like hallucination or toxicity detection, where simple binary judgments are often sufficient for assessing data quality.

\section{Tasks and Datasets}
\label{sec:3}

%The goal of our open-ended evaluation model is to assess the quality of a response according to human-defined criteria, producing both a final score and a natural language explanation for interpretability. Formally, given a task instruction $t$, input $i$, answer response(s) $a$, and an evaluation rubric $r$, the model is tasked to generate an explanation $e$ that justifies the evaluation and a corresponding score $s$ that reflects the response quality based on the evaluation rubric. We define this evaluation process as the following mapping:

The goal of our open-ended evaluation model is to assess the quality of a response according to human-defined criteria, producing both a final score and a natural language explanation for interpretability. Formally, given a task instruction $t$, input instance $i$, one or more candidate responses $a$, and an evaluation rubric $r$, the model is tasked with generating an explanation $e$, that justifies the evaluation and a score $s$ that reflects the response quality under the given rubric $r$. We define this evaluation process as a function:
\begin{equation}
f(x) = y, \quad \text{where } x = (t, i, a, r) \text{ and } y = (e, s).
\label{eq:main}
\end{equation}

\subsection{Task Formats}
\label{sec:3.1}

To support a wide range of evaluation settings, we define three task formats within our unified framework: \textit{point-wise}, \textit{pair-wise}, and \textit{binary} evaluation. Each format shares the same input structure $x = (t, i, a, r)$ and output structure $y = (e, s)$ but differs in how the candidate responses are structured and how the score $s$ is defined.

\paragraph{Point-wise Evaluation.}  
This format assesses the quality of a single response $a_1$ by assigning an integer score, typically on a 1–5 scale~\citep{kim2023prometheus}. It is suitable for open-ended generation tasks where scalar assessments of quality are needed, such as helpfulness, relevance, coherence, etc. Formally,
\begin{align}
a = a_1, \quad f_{point-wise}(t, i, a, r) = (e, s), \quad s \in \{1, 2, 3, 4, 5\}.
\end{align}

\paragraph{Pair-wise Evaluation.}  
In this setting, the model compares two candidate responses $a_1$ and $a_2$ to the same input $i$ and selects the preferred one, along with an explanation. This format is commonly used in preference-based training. Formally,
\begin{align}
a = (a_1, a_2), \quad f_{pair-wise}(t, i, a, r) = (e, s), \quad s \in \{a_1, a_2\}.
\end{align}

\paragraph{Binary Evaluation.}
Binary task requires the model to make a definitive judgment about the correctness or acceptability of a response $a_1$, given the input and rubric. These tasks span a variety of use cases, including factual verification, binary classification (e.g., determining whether a summary is faithful), and structured reasoning (e.g., assessing the validity of a math or code solution). Formally,
\begin{align}
a = a_1, \quad f_{binary}(t, i, a, r) = (e, s), \quad s \in \{\texttt{true}, \texttt{false}\}.
\end{align}
\begin{figure*}[!t]
    \centering
    \includegraphics[width=0.63\textwidth]{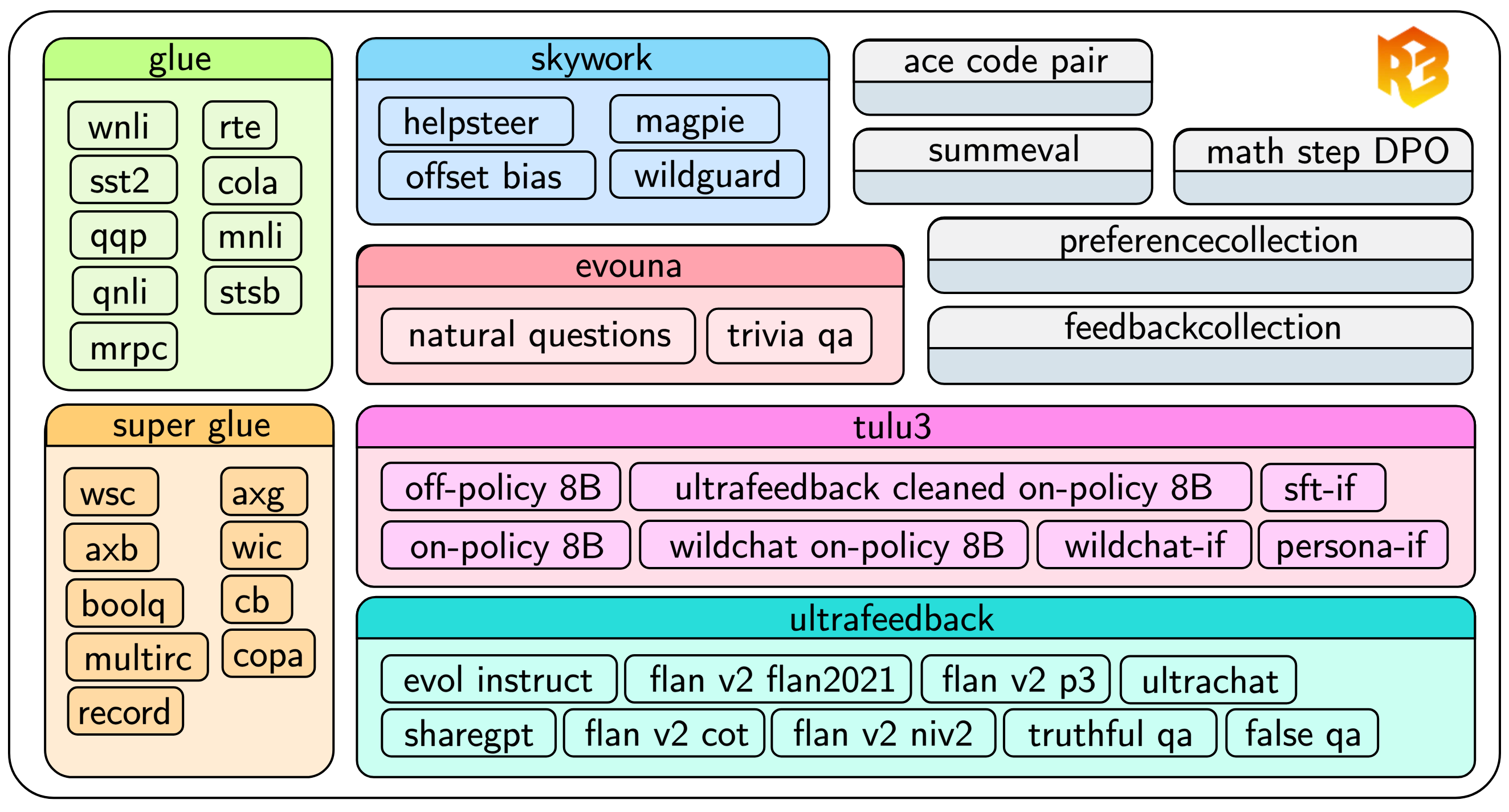}
    \caption{Dataset sources utilized in training the $\shortmethodname$ model.}
    \label{fig:datasets}
    \vspace{-1mm}
\end{figure*}

\subsection{\shortmethodname{} Datasets}
% label{sec:3.2}

To support open-ended evaluation across diverse domains and task formats, we begin with a large pool of publicly available datasets spanning over 4 million examples, which include general chat, reasoning, and classification tasks, as shown in Figure~\ref{fig:datasets}. However, most of these datasets lack consistent evaluation rubrics and explanation traces, which are key components to train our evaluation model to output both scores and natural language justifications. Generating such traces, particularly using a strong reasoning model such as DeepSeek-R1~\citep{guo2025deepseek}, is also computationally expensive and infeasible on a large scale.

To address this, we build our training dataset in multiple stages, drawing inspiration from \citet{muennighoff2025s1} to emphasize both quality and diversity of the training data while on a limited budget. We first sample a diverse subset from the raw pool, then enrich each example with on-the-fly rubric generation and explanation traces. Finally, we apply filtering and refinement to produce smaller, higher-quality datasets used in supervised training. The following sections describe each stage.

\subsubsection{Initial Curation}
We begin by curating a large collection of publicly available datasets, denoted by $\mathcal{D}_{init}$, which spans on three broad categories: general chat, reasoning, and classification or evaluation tasks. Each example \(x^{(j)} \in \mathcal{D}_{\mathrm{init}}\) is a tuple $x^{(j)} = \bigl(t^{(j)},\,i^{(j)},\,a^{(j)},\,r_{opt}^{(j)}\bigr)$, where $r_{opt}^{(j)}$ is optional rubric from the original dataset.

\begin{itemize}
\item \textbf{General Chat and Instruction-Following:}
This category includes open-domain instruction tuning and user preference data, drawn from resources such as the Tulu subset~\citep{lambert2024tulu3}, UltraFeedback~\citep{cui2023ultrafeedback}, and Skywork Reward Preference~\citep{liu2024skywork}. These datasets contain point-wise and pair-wise tasks.
\item \textbf{Reasoning Tasks:}  
To support math and code reasoning evaluations, we include datasets like Math-Step-DPO-10K~\citep{lai2024step} and AceCodePair-300K~\citep{zeng2025acecoder}, which contain preference annotations focused on correctness and reasoning quality on math and coding tasks.
\item \textbf{Classification and Factual Evaluation:}  
This category consists of binary and pair-wise tasks aimed at assessing factuality, consistency, and alignment with task rubrics. We include GLUE~\citep{wang2018glue}, SuperGLUE~\citep{wang2019superglue}, SummEval~\citep{fabbri2021summeval}, FeedbackCollection~\citep{kim2023prometheus}, PreferenceCollection~\citep{kim2024prometheus2}, and EVOUNA~\citep{wang2023evaluating}. These tasks span summarization, natural language inference, general rubric-based classification, and factual correctness.
\end{itemize}

To construct binary-labeled data that includes \emph{false} scores, we need to generate negative answers, as many datasets only provide the correct response (e.g., EVOUNA, GLUE, SuperGLUE). When possible, we sample negative answers from existing multiple-choice options. Otherwise, we generate negative answers using GPT-4.1 mini~\citep{achiam2023gpt}.

\subsubsection{Diversity Sampling}

To ensure feasibility for distilling reasoning traces while maintaining representative coverage across domains and reducing redundancy, we downsample \(\mathcal{D}_{\mathrm{init}}\) to a 20k-example subset \(\mathcal{D}_{20k} \subset \mathcal{D}_{\mathrm{init}}\), manually allocating quotas per dataset to balance task types and formats. For each dataset in $\mathcal{D}_{init}$, we perform a three-stage sampling process to extract the most diverse examples:

\begin{enumerate}
    \item \textbf{Embedding and Preprocessing.} Each instance is represented semantically by combining its instruction and input, $h(x^{(j)}) = t^{(j)} \oplus i^{(j)}$ and computing embeddings $Emb(h(x^{(j)})) = q^{(j)}.$

    \item \textbf{Cluster Determination and Assignment.} Samples are grouped into clusters, with the number of clusters $k^*$ chosen to maximize the average Silhouette score:
    \[
    k^* = \arg\max_{k} \frac{1}{|\mathcal{D}|} \sum_{j=1}^{|\mathcal{D}|} s_j^{(k)}, \quad
    s_j^{(k)} = \frac{v_j - w_j}{\max(v_j, w_j)},
    \]
    where $v_j$ and $w_j$ measure intra- and inter-cluster distances.  

    \item \textbf{Stratified Sampling with Maximal Marginal Relevance (MMR).} From each cluster, a subset is selected to balance topical relevance and diversity. For a candidate sample $x$, selected set $R$, and cluster $C$ with centroid $q_{C}$, the MMR score is
    \[
    \text{MMR}(x) = \lambda \cdot \text{sim}(x, q_C) - (1-\lambda) \cdot \max_{x_r \in R} \text{sim}(x, x_r),
    \]
    where $\lambda$ is a hyperparameter, and the next sample is chosen as $x^* = \arg\max_{x \in C \setminus R} \text{MMR}(x).$

\end{enumerate}

For binary datasets, we retain only one instance per question, either the positive or the negative, to avoid redundancy from semantically similar content. Further details on the final dataset composition and implementations are provided in Appendix Section~\ref{apx:detail-rm-train-datasets}.

\subsubsection{Rubric Generation}

Many datasets lack explicit evaluation rubrics, which are essential to our framework for generating structured supervision. To address this, we automatically generate rubrics based on task type at inference time. Formally, for each sample $x^{(j)}$ in $\mathcal{D}_{20k}$, we transform the optional rubric $r^{(j)}_{opt}$ into a required rubric $r^{(j)}$, so the dataset becomes $\mathcal{D}_{20k} = \{(t^{(j)},\,i^{(j)},\,a^{(j)},\,r^{(j)})\}_{j = 1}^{20000}$. The rubrics are generated based on task type using the following method:

\paragraph{Pair-wise and Binary Tasks.} We use templated prompts to generate rubric variations tailored to each format. To encourage generalization and mitigate overfitting, we randomize the rubric phrasing across three prompt variants. Full templates are listed in Appendices~\ref{apx:pair-wise-template} and~\ref{apx:binary-template}.

\paragraph{Point-wise Tasks.} When original rubrics $r_{opt}^{(j)}$ are available (e.g., in FeedbackCollection), we reuse them. Otherwise, we generate task-specific rubrics targeting relevant evaluation criteria (e.g., helpfulness in UltraFeedback) using a few-shot prompting strategy with GPT-4.1 mini. Details on rubric prompting are available in Appendix~\ref{apx:rubric-gen-template}.

Note that across all task types, rubric content is generated per instance and conditioned on the specific task and sample, so training and test splits would have their own rubrics depending on the specified tasks.

\subsubsection{Explanation Trace Generation}

Given the curated dataset \(\mathcal{D}_{20k} = \{(t^{(j)}, i^{(j)}, a^{(j)}, r^{(j)})\}_{j=1}^{20{,}000}\), we distill natural language explanations using a reasoning distillation model. Specifically, we define a function:
\begin{align}
\texttt{ReasoningModel}_\theta: (t^{(j)}, i^{(j)}, a^{(j)}, r^{(j)}) \longrightarrow \big(\text{reasoning\_trace}^{(j)},\, \hat{s}^{(j)},\, \hat{e}^{(j)}\big),
\end{align}
where \(\texttt{ReasoningModel}_\theta\) is instantiated with DeepSeek-R1~\citep{guo2025deepseek}. This model generates a natural language explanation (\(\text{reasoning\_trace}^{(j)}\)) along with short response of its predicted score \(\hat{s}^{(j)}\) and a short justification span \(\hat{e}^{(j)}\), following methodologies from prior work on explanation-based distillation~\citep{shridhar2023distilling, vu2024foundational}. Prompting templates are provided in Appendix~\ref{apx:prompt-templates}. We define the final target for each sample $x^{(j)}$ as:
\begin{equation}
y^{(j)} = \text{reasoning\_trace}^{(j)} \oplus (\hat{s}^{(j)}, \hat{e}^{(j)}),
\label{eq:target}
\end{equation}
where \(\oplus\) is string concatenation. Therefore, we define the dataset $\mathcal{D}_{20k}$ as \(\mathcal{D}_{20k} = \{(x^{(j)}, y^{(j)})\}_{j=1}^{20000}\).

We find that approximately 20\% of the reasoning traces are either overly verbose or contain repetitive content. For any example where \(y^{(j)}\) exceeds 4,096 tokens, we summarize the reasoning trace using GPT-4.1 mini. The summarization preserves the core explanation while removing redundant content and maintains stylistic coherence with the original output. Details and heuristics for this step are provided in Appendix~\ref{apx:trace-summarization}. 

As both the reasoning traces and their summaries are machine-generated, to verify the quality of the generated data, we conduct a human evaluation in Section~\ref{sec:human-eval-trace}, where we assess the factual correctness and logical coherence of the original reasoning traces, as well as the faithfulness and style consistency of the trace summarizations.

\subsubsection{Quality Filtering}

Finally, to improve the quality of our training dataset while preserving the diversity of responses, we apply a two-stage filtering pipeline to the annotated dataset \(\mathcal{D}_{20k}\).

\paragraph{Step 1: Incorrect Prediction Filtering.}
We discard examples for which the predicted score differs from the ground truth. Formally, we construct a filtered dataset \(\mathcal{D}_{14k} \subset \mathcal{D}_{20k}\) such that for each retained example \((x^{(j)}, y^{(j)}) \in \mathcal{D}_{14k}\), we have $\hat{s}^{(j)} = s^{(j)}$, where $s^{(j)}$ is the true score for sample $x^{(j)}$.
This ensures that all reasoning signals used for training are consistent with the gold labels. After this step, approximately 14,000 examples remain.

\paragraph{Step 2: Triviality Filtering via Small Model Agreement.}
To remove overly easy examples that provide a limited training signal, we evaluate each example in \(\mathcal{D}_{14k}\) using our smallest model, Qwen3-4B~\citep{yang2025qwen3}. For each example $x^{(j)}$, we compute predictions across five decoding runs without chain-of-thought reasoning as $\{\hat{s}^{(j)}_{[1]}, \dots, \hat{s}^{(j)}_{[5]}\} = \texttt{Qwen3-4B}(x^{(j)})$. If $\hat{s}^{(j)}_{[k]} = s^{(j)}$ for all $k \in \{1,\dots,5\}$, then we discard $x^{(j)}$
This results in the final dataset \(\mathcal{D}_{4k} \subset \mathcal{D}_{14k}\), containing approximately 4,000 challenging and diverse training examples. We fine-tune our $\shortmethodname$ models with both $\mathcal{D}_{4k}$ (\textsc{-4k}) and $\mathcal{D}_{14k}$ (\textsc{-14k}) to assess the impact of data size. Detailed dataset statistics are provided in Appendix Section~\ref{apx:detail-rm-train-datasets}.

\subsection{Reward Models Training and Evaluation}
\label{sec:3.3}
Given our generated training data, we further use supervised fine-tuning (SFT) to enhance the base model's reasoning capability as a reward model by minimizing the negative log-likelihood of reference responses.  Given our training dataset \(\mathcal{D} = \{(x^{(i)}, y^{(i)})\}_{i=1}^N\), where \(x^{(i)}\) is prompt input previously introduced in eq. (\ref{eq:main}) and \(y^{(i)} = (y^{(i)}_1, \dots, y^{(i)}_{T_i})\) introduced in eq. (\ref{eq:target}) is the corresponding target sequence, the objective is the cross-entropy loss:
\begin{align}
\mathcal{L}_{\mathrm{SFT}}(\theta)
= -\frac{1}{N} \sum_{i=1}^N \sum_{t=1}^{T_i}
\log \; \pi_\theta\!\bigl(y^{(i)}_t \mid y^{(i)}_{<t},\, x^{(i)}\bigr)\,,
\end{align}
where \(\pi_\theta(y_t \mid y_{<t}, x)\) denotes the model’s conditional probability of token \(y_t\) given the history \(y_{<t}\) and prompt \(x\), parameterized by \(\theta\).  By directly maximizing the log-likelihood of the ground-truth tokens, this loss encourages the base model to produce high-quality reasoning traces and the desired format for pair-wise comparisons or single-answer rewards. We further investigate lightweight fine-tuning via Low-rank Adaptation (LoRA)~\citep{hu2022lora} on \textsc{$\shortmethodname$-4K} data as part of our additional study to reduce training costs and data requirements while maintaining competitive performance.

Given that our tasks are within multiple domains, including but not limited to verifiable tasks like math and coding, reinforcement learning (RL) techniques are not directly applicable here. Nevertheless, it is possible to explore the effectiveness of using our trained model as a reward signal for enhancing other models' capabilities via RL. However, as shown in our ablation study in Section~\ref{sec:app-sft-vs-rl} in terms of computational cost and performance benefits, we opted for SFT instead.

For our $\shortmethodname$ models, we mainly perform SFT on the Qwen3 model family~\citep{yang2025qwen3} on the 4B, 8B, and 14B scales, along with Phi-4-reasoning-plus~\citep{abdin2025phi}. For fair comparison with other baselines, we also perform SFT on Qwen2.5-Instruct-7B~\citep{team2024qwen2}, DeepSeek-Distilled-Qwen-14B~\citep{guo2025deepseek}. For evaluation, we compare our open-source models against both open-source and proprietary baselines. Among open-source baselines, we primarily evaluate against \texttt{Prometheus-7B-v2.0}~\citep{kim2024prometheus2}, a popular rubric-based LLM-as-a-judge, and RM-R1~\citep{chen2025rm} model suites, the most recent reasoning reward model work at the time of writing. For proprietary comparisons, we include DeepSeek-R1~\citep{guo2025deepseek}, GPT-4.1 mini, GPT-o4 mini, and GPT-5 mini, strong closed models recognized for their general and reasoning-specific capabilities. Additionally, we include models that have reported results in their respective publications.

Finally, we evaluate the reward models across a diverse suite of benchmarks spanning multiple evaluation paradigms. For pairwise preference scoring, we use RM-Bench~\citep{liu2024rm} and RewardBench~\citep{lambert2024rewardbench}. For pointwise scoring, we employ XSUM~\citep{narayan2018don} and FeedbackBench~\citep{kim2023prometheus}. For binary classification tasks, we adopt MMLU-STEM~\citep{hendrycks2021mmlu} and BBH~\citep{suzgun2022bbh}. All experiments are run across 5 different seeds. More details about the evaluation datasets and how we constructed them for evaluation can be found in Appendix Section~\ref{apx:detail-rm-eval-datasets}.

\subsection{Policy Model Alignment Training and Evaluation}
\label{sec:align-eval}

For policy model alignment, we use LLaMA 3.2–3B Instruct as the base policy model and perform Direct Preference Optimization (DPO)~\citep{rafailov2023direct} on the HelpSteer3-Preference dataset~\citep{wang2025helpsteerpreference}. We focus on DPO as RM-R1 is only applicable to preference data. In addition, to reduce confounding factors, we restrict training to the English portion of the dataset and to single-turn interactions, which ensures consistency with our evaluation setup. We also provide REINFORCE++ and GRPO experiments for on-policy optimization in Section~\ref{sec:on-policy-opt-exps}, as the other baseline reward models are restricted to pairwise scoring and therefore cannot be used for point-wise on-policy optimization.

We evaluate the aligned policy models using two widely adopted benchmarks: MT-Bench~\citep{zheng2023judging} and WildBench~\citep{lin2024wildbench}, following the evaluation protocol of~\citet{wang2025helpsteer3}. We do not use AlpacaEval2~\citep{dubois2024length}, since WildBench contains more challenging and realistic prompts compared to AlpacaEval2~\citep{wang2025helpsteer3}. In both cases, model outputs are judged by GPT-4.1-mini, which we select over GPT-4o-Turbo due to its stronger empirical reliability as an evaluator and being cheaper. This setup allows us to measure the effectiveness of preference-based alignment in improving model helpfulness, robustness, and general instruction-following capability.

% \begin{table*}[!th]
% \centering
% \caption{Ablation studies on dataset construction, employing the R3\textsc{-Qwen3-14B} model trained on a 14k-sample dataset using LoRA.}
% \resizebox{\textwidth}{!}{
%     \begin{tabular}{lcccccccc}
%     \toprule
%     \textbf{} & \multicolumn{1}{c}{RM-Bench} & \multicolumn{1}{c}{RewardBench} & \multicolumn{1}{c}{BBH} & \multicolumn{1}{c}{MMLU-STEM} & \multicolumn{3}{c}{XSUM} & \multicolumn{1}{c}{FeedbackBench} \\
%     & Overall Acc. & Overall Acc. & Overall Acc. & Overall Acc. & Acc. Faithfulness & $\tau$ Coherence & $\tau$ Relevance & $\tau$ \\ 
%     \midrule
%     Random Sampling & 77.0 & 86.6 & 89.7 & 93.0 & 71.0 & 0.228 & 0.333 & 0.648 \\ \midrule
%     Dataset  & & & & && & & \\
%     $\quad$Only Pairwise & 82.1 & \textbf{90.2} & 91.5 & 94.4 & 68.4 & 0.205 & 0.329 & 0.662 \\
%     $\quad$Only Pointwise & 80.0 & 86.0 & 90.1 & 93.4 & 63.5 & 0.236 & 0.318 & 0.639 \\
%     $\quad$Only Binary & 81.6 & 88.8 & 91.0 & 94.0 & 67.2 & \textbf{0.296} & 0.332 & 0.658 \\ \midrule
%     No Rubric & 76.3 & 87.9 & 85.1 & 91.9 & 70.3 & -0.037 & -0.032 & 0.652 \\ 
%     No Explanation & 83.1 & \textbf{90.2} & 91.7 & \textbf{94.5} & 66.4 & 0.198 & 0.353 & 0.651 \\
%     No Reasoning & 71.2 & 82.6 & 79.8 & 88.2 & \textbf{72.8} & 0.092 & \textbf{0.399} & \textbf{0.683} \\ \midrule
%     $\shortmethodname$ & \textbf{83.5} & \textbf{90.2} & \textbf{91.9} & \textbf{94.5} & 69.2 &	0.245 & 0.336 & 0.654 \\
%     \bottomrule
%     \end{tabular}
% }
% \label{results-ablation}
% \vspace{-3mm}
% \end{table*}

\begin{table*}[!t]
\centering
\caption{Comparison of existing models with $\shortmethodname$ trained with 14K samples on RM-Bench. \textbf{Bolded numbers} indicate the best-performing results between $\shortmethodname$ models and baseline models. Proprietary models are bolded and compared independently.}
\resizebox{0.82\textwidth}{!}{
    \begin{tabular}{lcccc|ccc|c}
    \toprule
    \textbf{Model} & \multicolumn{4}{c|}{Domain} & \multicolumn{3}{c|}{Difficulty} & Overall \\
    & Chat & Math & Code & Safety & Easy & Medium & Hard & Avg. \\ 
    \midrule
    Base Models \\ \midrule
    Qwen2.5-7B & 59.9 & 52.6 & 50.9 & 72.4 & 66.4 & 60.3 & 50.1 & 58.9 \\
    DeepSeek-Distilled-Qwen-14B & 65.0 & 85.4	& 69.0 & 85.1 & 84.3	& 78.7	& 65.4 & 76.1 \\
    Qwen3-4B & 64.3	& 84.0	& 62.0	& 85.6	& 82.6	& 74.5	& 64.8	& 74.0 \\
    Qwen3-8B&  63.2	& 80.5	& 61.0	& 84.8	& 80.1	& 73.4	& 63.5	& 72.4 \\
    Qwen3-14B & \textbf{76.4} & 88.0	& 70.7 & 91.0 & 87.9 & 84.3 & 72.3 & 81.5 \\
    Phi-4-R$^+$-14B & 73.4 & 87.3 & 70.7 &	85.7 & 86.5 & 80.2 & 71.1 &	79.3 \\
    \midrule
    Existing Reward Models \\ \midrule
    % ArmoRM & ? & ? & ? & ? & ? & ? & ? & ?\\
    Prometheus-7B-v2.0 & 46.0 & 52.6 & 47.6 & 73.9 & 68.8 & 54.9 & 41.3 & 55.0 \\ 
    % \midrule
    % Qwen3 4B & 67.7 & 90.1 & 70.4 & 86.4 & 87.9 & 79.3 & 68.8 & 78.7 \\
    % Qwen3 8B & 69.2 & 89.2 & 71.4 & 87.0 & 86.6 & 81.0 & 70.0 & 79.2 \\
    % Qwen3 14B & &  & & & & & & \\
    JudgeLRM & 59.9 & 59.9 & 51.9 & 87.3 & 73.2 & 76.6 & 54.8 & 64.7 \\
    RM-R1-Qwen-Instruct-7B & 66.6 & 67.0 & 54.6 & 92.6 & 79.2 & 71.7 & 59.7 & 70.2 \\
    RM-R1-DeepSeek-Distilled-Qwen-7B & 64.0 & 83.9 & 56.2 & 85.3 & 75.9 & 73.1 & 68.1 & 72.4 \\
    RM-R1-Qwen-Instruct-14B & 75.6 & 75.4 & 60.6 & 93.6 & 82.6 & 77.5 & 68.8 & 76.1 \\
    RM-R1-Qwen-Instruct-32B & 75.3 & 80.2 & 66.8 & 93.9 & 86.3 & 80.5 & 70.4 & 79.1 \\
    RM-R1-DeepSeek-Distilled-Qwen-14B & 71.8 & 90.5 & 69.5 & 94.1 & 86.2 & 83.6 & 74.4 & 81.5 \\
    RM-R1-DeepSeek-Distilled-Qwen-32B & 74.2 & 91.8 & 74.1 & \textbf{95.4} & 89.5 & 85.4 & \textbf{76.7} & 83.9 \\ 
    \midrule
    $\shortmethodname$ Models (Ours) \\ \midrule
    $\shortmethodname\textsc{-Qwen2.5-7B}$ &  66.8 & 82.0 & 65.0 & 87.0 & 83.8 & 76.8 & 64.9 & 75.2 \\
    $\shortmethodname\textsc{-DeepSeek-Distilled-Qwen-14B}$ & 71.7 & 93.0 & 78.4 & 86.4 & 89.3 & 84.7 & 73.1 & 82.4 \\
    $\shortmethodname\textsc{-Qwen3-4B}$ & 67.9 & 93.0 & 74.7 & 86.9 & 88.8 & 81.9 & 71.2 & 80.6  \\
    $\shortmethodname\textsc{-Qwen3-8B}$ & 69.1 & 93.2 & 75.9 & 87.6 & 89.0 & 83.4 & 71.9 & 81.4 \\
    $\shortmethodname$\textsc{-Qwen3-14B} & 73.4 & \textbf{93.8} & \textbf{79.1} & 89.5 & \textbf{90.3} & \textbf{86.6} & 74.9 & \textbf{84.0} \\ 
    $\shortmethodname$\textsc{-Phi-4-R$^+$-14B} & 74.5 & 93.0 & 77.5 & 84.8 & 89.3 & 84.7 & 73.3 & 82.5 \\ \midrule
    Proprietary Models \\ \midrule
    GPT-4.1 mini & 67.6 & 73.0 & 71.3 & 90.7 & 87.0 & 78.4 & 61.7 & 75.7 \\
    GPT-o4 mini & 77.6 & \textbf{93.0} & 80.8 & \textbf{93.4} & \textbf{92.0} & \textbf{88.7} & 78.0 & 86.2 \\
    GPT-5 mini & \textbf{88.0} & 92.9 & \textbf{91.1} & 78.0 & 77.4 & 85.8 & \textbf{96.4} & \textbf{92.4} \\
    DeepSeek-R1 & 78.6 & 66.2 & 81.9 & 88.7 & 86.9 & 82.2 & 67.3 & 78.8 \\
    \bottomrule
    \end{tabular}
}
\label{results-rmbench}
\vspace{-2mm}
\end{table*}

\section{Results and Analysis}
\label{sec:4}

In this section, we present the overall performance of the reward modeling results. We mainly show results of $\shortmethodname$ models trained with 14K samples. More detailed results, including results of our models that are trained using different strategies and results of other baseline models, can be found in the Appendix Section~\ref{apx:detailed-results-break}.

\begin{table*}[!th]
\centering
\caption{Comparison of existing models with $\shortmethodname$ on RewardBench using pair-wise scoring. \textbf{Bolded numbers} indicate the best-performing results between $\shortmethodname$ models and baseline models. Proprietary models are bolded and compared independently.}
\resizebox{0.78\textwidth}{!}{
    \begin{tabular}{lcccc|c}
    \toprule
    \textbf{Models} & Chat & Chat Hard & Safety & Reasoning & Avg. \\ 
    \midrule
    Base Models \\ \midrule
    Qwen2.5-7B & 86.3 & 60.1 & 77.8 & 72.1 & 74.1 \\
    DeepSeek-Distilled-Qwen-14B & 91.1 & 72.8 & 83.0 & 88.5 & 83.8 \\
    Qwen3-4B & 88.6 & 65.1 & 83.8 &	89.7 & 81.8 \\
    Qwen3-8B& 89.1 & 70.0 & 82.7 & 88.9 & 82.7 \\
    Qwen3-14B & 93.6 & 78.5 & 87.7 & 95.1 & 88.7 \\
    Phi-4-R$^+$-14B & 92.7 & 79.6 & 85.9 & 90.1 & 87.1 \\
    \midrule
    Existing Reward Models \\ \midrule
    % ArmoRM & 96.9 & 76.8 & 90.5 & 97.3 & 90.4\\
    Prometheus-7B-v2.0 & 90.2 & 45.6 & 75.8 & 74.6 & 71.6\\ 
    m-Prometheus-14B & 93.6 & 59.0 & 85.1 & 84.8 & 80.6 \\ 
    JudgeLRM & 92.9 & 56.4 & 78.2 & 73.6 & 75.2 \\
    SynRM & 38.0 & 82.5 & 74.1 & 87.1 & 70.4\\
    % \midrule
    % Reasoning Models \\ \midrule
    % Qwen3 4B & 83.5 & 62.3 & 79.6 & 71.9 & 74.3 \\
    % Qwen3 8B & 93.0 & 68.0 & 81.9 & 92.9 & 83.9 \\
    % Qwen3 14B & 92.2 & 74.6 & 83.8 & 93.6 & 86.0 \\
    % RM-R1 7B & 94.1 & 74.6 & 85.2 & 86.7 & 85.2 \\
    % RM-R1 14B & 91.3 & 79.4 & 89.3 & 95.5 & \textbf{88.9}\\
    RM-R1-DeepSeek-Distilled-Qwen-7B & 88.9 & 66.2 & 78.4 & 87.0 & 80.1 \\
    RM-R1-Qwen-Instruct-7B & 94.1 & 74.6 & 85.2 & 86.7 & 85.2 \\
    RM-R1-Qwen-Instruct-14B & 93.6 & 80.5 & 86.9 & 92.0 & 88.2 \\
    RM-R1-DeepSeek-Distilled-Qwen-14B & 91.3 & 79.4 & 89.3 & 95.5 & 88.9 \\
    RM-R1-DeepSeek-Distilled-Qwen-32B & 95.3 & 80.3 & 91.1 & 96.8 & 90.9 \\
    RM-R1-Qwen-Instruct-32B & \textbf{95.3} & \textbf{83.1} & \textbf{91.9} & 95.2 & \textbf{91.4} \\
    \midrule
    $\shortmethodname$ Models (Ours) \\ \midrule
    $\shortmethodname\textsc{-Qwen2.5-7B}$ & 91.4 & 73.8 & 85.1 & 90.6 & 85.2 \\
    $\shortmethodname\textsc{-DeepSeek-Distilled-Qwen-14B}$ & 92.3 & 77.8 & 86.8 & 95.6 & 88.1 \\
    $\shortmethodname\textsc{-Qwen3-4B}$ & 92.4	& 76.0	& 85.8	& 95.7 & 87.5	\\
    $\shortmethodname\textsc{-Qwen3-8B}$ & 93.8	&  78.6	& 86.3	& 96.7 & 88.8	\\
    $\shortmethodname\textsc{-Qwen3-14B}$ & 93.3 & 79.7 & 88.4	& \textbf{96.9} & 89.6	\\
    $\shortmethodname$\textsc{-Phi-4-R$^+$-14B} & 94.5 & 78.0 & 86.6 & 96.5 & 88.9 \\\midrule
    Proprietary Models \\ \midrule
    GPT-4.1 mini & \textbf{96.1} & 75.2 & 87.0 & 89.6 & 87.0 \\ 
    GPT-o4 mini & 95.3 & \textbf{81.8} & 91.6 & \textbf{98.4} & \textbf{91.8} \\
    GPT-5 mini & 95.3 & 81.6 & \textbf{92.0} & \textbf{98.4} & \textbf{91.8} \\
    DeepSeek-R1 & 93.6 & 79.2 & 86.9 & 97.4 & 89.3 \\
    \bottomrule
    \end{tabular}
    }
\label{all-results-rewardbench}
\end{table*}

\subsection{Reward Models Evaluation}

\subsubsection{Overall Performance}
Table~\ref{results-rmbench} and Table~\ref{all-results-rewardbench} highlight the strong performance of our $\shortmethodname{}$ models on RM-Bench and RewardBench, showcasing the effectiveness of $\shortmethodname{}$ models in pair-wise preference scoring under a training budget. Particularly in RM-bench, our models deliver remarkable results where even our smallest model, \shortmethodname\textsc{-Qwen3-4B}, outperforms nearly all other reasoning models from RM-R1, with the exception of RM-R1-DeepSeek-Distilled-Qwen-14B and RM-R1-DeepSeek-Distilled-Qwen-32B. It also surpasses Prometheus-7B-v2.0, GPT-4.1 mini, and even DeepSeek-R1 as well, demonstrating its competitiveness. When comparing the same base models with RM-R1, our models \shortmethodname\textsc{-Qwen2.5-7B} and \shortmethodname\textsc{-DeepSeek-Distilled-Qwen-14B} outperform their RM-R1 counterparts up to 6.3 points. 

Other variants of our $\shortmethodname$ models also show competitive performance. For instance, \shortmethodname\textsc{-Qwen3-14B-LoRA-4k} model (shown in the Appendix on Table~\ref{tab:results-rmbench-supplementary}) achieves the state-of-the-art performance, despite being trained with LoRA and smaller data. Between our $\shortmethodname$\textsc{-Qwen3-14B} and $\shortmethodname$\textsc{-Phi-4-R$^+$} models, $\shortmethodname$\textsc{-Qwen3-14B} models consistently outperforms $\shortmethodname$\textsc{-Phi-4-$R^+$} models in all aspects. Overall, the exceptional performance of our $\shortmethodname$ models highlights the robustness and effectiveness of our training approach and meticulous data curation strategies.

In RewardBench, we achieve similarly impressive results. Our \shortmethodname\textsc{-Qwen3-4B} models, despite being half the size of the RM-R1 7B models, outperform all RM-R1 7B models as well as Prometheus-7B-v2.0 by at least 1.8 points. Furthermore, the \shortmethodname\textsc{-Qwen3-4B-14k} model surpasses GPT-4.1 mini by 0.5 points. When assessing our \shortmethodname\textsc{-Qwen3-14B} models against the RM-R1 14B model families, the \shortmethodname\textsc{-Qwen3-14B-LoRA-4k} model (shown in the Appendix on Table~\ref{tab:results-rewardbench-supplementary}) surpasses RM-R1-DeepSeek-Distilled-Qwen-14B by 0.4 points, while matching the average performance of DeepSeek-R1. Between our $\shortmethodname$\textsc{-Qwen3-14B} and $\shortmethodname$\textsc{-Phi-4-R$^+$} models, the $\shortmethodname$\textsc{-Qwen3-14B} models generally outperform the $\shortmethodname$\textsc{-Phi-4-R$^+$} models, except in the Chat and Safety categories. Notably, our models demonstrate competitive performance even compared with DeepSeek-R1 under training budget constraints in terms of data and memory.

Table~\ref{tab:point-wise-results} presents the performance of our $\shortmethodname{}$ models on point-wise assessment tasks, which are XSUM and FeedbackBench, along with binary tasks from BBH and MMLU-STEM. For XSUM, all $\shortmethodname{}$ models consistently outperform DeepSeek-R1 and Prometheus-7B-v2.0 in terms of faithfulness, highlighting the effectiveness of binary assessment for $\shortmethodname{}$ models. In terms of coherence and relevance, our $\shortmethodname$\textsc{-Phi-4-R$^+$} models perform the best among all open-source and proprietary models. On the generic rubric-based assessment from FeedbackBench, $\shortmethodname{}$ models perform competitively against all baselines. While Prometheus-7B-v2.0 achieves the highest score on FeedbackBench, its relatively weaker performance on other benchmarks suggests that it may be better aligned with the specific rubric or distribution of FeedbackBench, rather than generalizing across diverse tasks.

For binary classification tasks such as BBH and MMLU-STEM, we observe that both larger model size and greater fine-tuning data improve performance, reflecting stronger reasoning capabilities. All of our $\shortmethodname{}$ models outperform Prometheus-7B-v2.0, while \shortmethodname\textsc{-Qwen3-14B} models surpass GPT-4.1 mini's performance. Overall, these results highlight the competitive and robust performance of $\shortmethodname{}$ models across a range of point-wise and binary evaluation tasks.

\begin{figure*}[!th]
    \centering
    \includegraphics[width=1.0\linewidth]{assets/performance.png}
    \caption{$\shortmethodname{}$ models outperforms competitor models across differences model sizes in all data types.}
    \label{fig:overall-performance}
\end{figure*}

\begin{table*}[!th]
\centering
\caption{Comparison of existing models with $\shortmethodname{}$ on XSUM, FeedbackBench, BBH Binary, and MMLU-STEM Binary. \textbf{Bolded numbers} indicate the best-performing results between $\shortmethodname$ models and baseline models. Proprietary models are bolded and compared independently.}
\label{tab:point-wise-results}
\resizebox{\textwidth}{!}{
    \begin{tabular}{lccc|c|c|c}
    \toprule
    \textbf{Models} & \multicolumn{3}{c|}{\textbf{XSUM}} & \textbf{FeedbackBench} & \textbf{BBH Binary} & \textbf{MMLU-STEM Binary}\\ 
    & Acc. &  \multicolumn{2}{c|}{Kendall Tau} & Kendall Tau & Acc. & Acc. \\ 
    & Faithfulness & Coherence & Relevance & &  &  \\
    \midrule
    Base Models \\ \midrule
    Qwen2.5-7B & 57.3 & -0.04 & 0.30 & 0.66 & 59.9 & 69.6 \\
    DeepSeek-Distilled-Qwen-14B & 65.0 & 0.26 & 0.28 & 0.63 &87.5 & 90.7 \\
    Qwen3-4B & \textbf{75.9} &0.17 & 0.25 & 0.67  & 89.0 & 91.1 \\
    Qwen3-8B& 72.5 &0.28 &0.33 & 0.67  & 90.9 &  92.4 \\
    Qwen3-14B & 70.2 &0.31 &0.34 & 0.67  & 91.8 & 93.2 \\
    Phi-4-R$^+$-14B & 64.0 & 0.46 &0.30 & 0.70  & 90.8 &  92.9\\
    \midrule
    Existing Reward Models \\ \midrule
    Prometheus-7B-v2.0 & 60.7 & 0.12 & 0.16 & \textbf{0.79} & 54.0 & 56.5 \\ 
    Selene-1-Mini-Llama-3.1-8B & 56.4 & 0.16 & 0.36 & 0.78 & 58.2 & 65.2\\
    RISE-Judge-Qwen2.5-7B & 66.4 & 0.29 & 0.32 & 0.68 & 63.1 & 76.9 \\
    RISE-Judge-Qwen2.5-32B & 
    71.0 & 0.30 & \textbf{0.39} & 0.74 & 82.8 & 89.4 \\ \midrule
    $\shortmethodname$ Models (Ours) \\ \midrule
    $\shortmethodname\textsc{-Qwen2.5-7B}$ & 67.5 & 0.33 & 0.35 & 0.70 & 81.1 & 88.3 \\
    $\shortmethodname\textsc{-DeepSeek-Distilled-Qwen-14B}$ & 64.3 & \textbf{0.41} & 0.34 & 0.72 & 91.1 & 93.0 \\
    $\shortmethodname\textsc{-Qwen3-4B}$ & 66.7	& 0.26 & 0.31 & 0.68 & 89.3 & 92.0 \\
    $\shortmethodname\textsc{-Qwen3-8B}$ & 65.8 & 0.37 & 0.33 & 0.71 & 90.7 & 93.6 \\
    $\shortmethodname\textsc{-Qwen3-14B}$ & 68.5 & 0.33 & 0.37 & 0.72 & 92.1 & \textbf{94.8} \\
    $\shortmethodname$\textsc{-Phi-4-R$^+$-14B} & 67.3 & 0.36 & 0.34 & 0.71 & \textbf{92.2} & 94.4 \\ \midrule
    Proprietary Models \\ \midrule
    GPT-4.1 mini & \textbf{72.6} & 0.07 & \textbf{0.38} & 0.69 & 91.0 & 93.3 \\
    GPT-o4 mini & 69.1 & 0.16 & 0.30 & 0.65 & 93.2 & 95.3\\
    GPT-5 mini & 68.7 & \textbf{0.42} & \textbf{0.39} & 0.62 & \textbf{95.0} & \textbf{96.5} \\
    DeepSeek-R1 & 60.4 & 0.35 & 0.38 & \textbf{0.72} & 94.0 & 96.2 \\
    \bottomrule
    \end{tabular}
    }
\end{table*}

\subsubsection{Model Scaling and Efficiency}

Figure~\ref{fig:overall-performance} presents averaged results under three evaluation settings: binary, pairwise, and pointwise. Larger models generally demonstrate stronger reasoning and reward generation capabilities, and our $\shortmethodname$ models show consistent improvements as parameter size increases, with some benchmarks exhibiting substantial gains. Within the same parameter size, $\shortmethodname$ models achieve the best performance. Moreover, smaller variants of our $\shortmethodname$ models can sometimes outperform larger models. For example, $\shortmethodname\textsc{-Qwen3-4B}$ outperforms Rise-judge-qwen2.5-32B and Prometheus-7B-v2.0 in the binary setting, while $\shortmethodname\textsc{-Qwen3-14B}$ outperforms RM-R1 32B in the pairwise setting. These results suggest that both our methodology and dataset are highly effective for training reward models in resource-constrained settings.

\subsubsection{Generalization}

Among proprietary models, GPT-4o-mini outperforms DeepSeek-R1 on reward benchmarks involving pair-wise scoring, while DeepSeek-R1 demonstrates stronger performance on tasks such as XSUM, FeedbackBench, BBH, and MMLU-STEM. For open-weight models, our $\shortmethodname$ models consistently outperform existing reward models, such as Prometheus-7B-v2.0 and all RM-R1 variants, across most benchmarks. The only exception is FeedbackBench, where Prometheus-7B-v2.0 performs exceptionally well. However, this suggests that Prometheus-7B-v2.0 is highly specialized rather than robust across tasks. In contrast, RM-R1 is more robust than Prometheus-7B-v2.0 but lacks flexibility in supporting diverse evaluation formats such as point-wise and binary scoring; Prometheus, meanwhile, supports only point-wise and pair-wise formats. Our $\shortmethodname$ models offer both robustness and versatility, making it more suitable for general-purpose reward modeling. Because rubric content and prompts are instance-specific and do not overlap between training and test splits, every test instance inherently requires generalization to a previously unseen rubric or prompts, so there is no ``seen-rubric'' subset to serve as a contrastive baseline.

\subsubsection{Ablation Studies}

\begin{table*}[!th]
\centering
\caption{Ablation studies on dataset construction, employing the $\shortmethodname$\textsc{-Qwen3-14B} model trained on a 14k-sample dataset using LoRA. \textbf{Bolded numbers indicate the best-performing results.}}
\resizebox{0.7\textwidth}{!}{
    \begin{tabular}{lcccc}
    \toprule
    \textbf{} & \multicolumn{1}{c}{RM-Bench} & \multicolumn{1}{c}{RewardBench} & \multicolumn{1}{c}{BBH} & \multicolumn{1}{c}{MMLU-STEM} \\
    & Overall Acc. & Overall Acc. & Overall Acc. & Overall Acc. \\ 
    \midrule
    Random Sampling & 77.0 & 86.6 & 89.7 & 93.0  \\ \midrule
    Dataset  & & &\\
    $\quad$Only Pairwise & \textbf{82.1} & \textbf{90.2} & \textbf{91.5} & \textbf{94.4} \\
    $\quad$Only Pointwise & 80.0 & 86.0 & 90.1 & 93.4 \\
    $\quad$Only Binary & 81.6 & 88.8 & 91.0 & 94.0 \\ \midrule
    No Rubric & 76.3 & 87.9 & 85.1 & 91.9  \\ 
    No Explanation & \textbf{83.1} & \textbf{90.2} & \textbf{91.7} & \textbf{94.5}\\
    No Reasoning & 71.2 & 82.6 & 79.8 & 88.2  \\ \midrule
    $\shortmethodname$ & \textbf{83.5} & \textbf{90.2} & \textbf{91.9} & \textbf{94.5} \\
    \bottomrule
    \end{tabular}
}
\label{results-ablation}
\vspace{-2mm}
\end{table*}

We conduct an ablation study to assess the effectiveness of our overall dataset construction on different sampling strategies, dataset types, and supervision signals, with results summarized in Table~\ref{results-ablation}. For efficiency, we apply LoRA~\citep{hu2022lora} in all experiments using $\shortmethodname$\textsc{-Qwen3-14B}. Our results indicate that random sampling consistently underperforms compared to diversity sampling. Among dataset types, pairwise supervision achieves the best results (82.1\% on RM-Bench, 94.4\% on MMLU-STEM), surpassing pointwise and binary-only settings and improving relevance on XSUM. Supervision signals also have distinct effects: removing the rubric lowers BBH accuracy, excluding explanations reduces coherence, and eliminating reasoning traces causes the largest performance drop (e.g., 71.2\% RM-Bench, 79.8\% BBH), underscoring the importance of reasoning data. The full model ($\shortmethodname$) achieves the best overall balance (83.5\% RM-Bench, 94.5\% MMLU-STEM, strong scores on coherence, relevance, and FeedbackBench). Although excluding explanations has a limited impact on accuracy, we retain them in $\shortmethodname$ to enable more interpretable outputs.

\begin{table*}[!t]
\centering
\caption{$\shortmethodname$ Preference Optimization results using GPT-4.1-mini as the judge model. Our $\shortmethodname$ models outperform a larger reward model, LLaMA-3.3-Nemotron-Super-49B-GenRM-Multilingual, despite the latter having over three times more parameters. \textbf{Bolded numbers} indicate the best performing results, while \underline{underlined numbers} indicate the second-best performing results.}
\resizebox{\textwidth}{!}{
    \begin{tabular}{lcc|ccccccc}
    \toprule
    Method & RM \#Params & MT-Bench & & & WildBench \\ 
    & & Overall & Overall & Creative Plan. & Data Analy. & Info. & Seek. & Coding \\ \midrule
    LLaMA 3.2 3B Instruct (Init Policy) & & 5.75 & 15.70 & 43.10 & 25.12 & 3.65 & 34.44 & -5.97 \\
    \midrule
    + DPO w/ RM-R1-DeepSeek-Distilled-Qwen-14B & 14B & 5.98 & 20.42 & 50.28 & 29.39 & 8.41 & 41.29 & -3.41 \\
    + DPO w/ RM-R1-DeepSeek-Distilled-Qwen-32B & 32B & 6.01 & 22.28 & \textbf{54.32} & \underline{31.66} & 9.37 & 43.91 & -2.56 \\
    + DPO w/ LLaMA-3.3-Nemotron-Super-GenRM & 49B & \textbf{6.38} & 22.83 & \underline{54.06} & 31.33 & 10.12 & \textbf{44.91} & -1.24 \\
    + DPO w/ LLaMA-3.3-Nemotron-Super-GenRM-Multilingual & 49B & \underline{6.20} & \underline{23.30} & 52.71 & \textbf{31.93} & \underline{11.43} & \underline{44.90} & \underline{-0.57} \\
    \midrule
    $\shortmethodname$ Models (Ours) \\
    \midrule
    + DPO w/ $\shortmethodname$\textsc{-Qwen3-4B-14k} & 4B & 5.95 & 22.00 & 51.83 & 30.78 & 9.08 & 43.42 & -1.23 \\
    + DPO w/ $\shortmethodname$\textsc{-Qwen3-8B-14k} & 8B & 6.13 & 21.67 & 52.05 & 29.94 & 9.44 & 41.49 & -0.85 \\
    + DPO w/ $\shortmethodname$\textsc{-Qwen3-14B-LoRA-4k} & 14B & \underline{6.20} & \textbf{23.45} & 52.57 & 31.62 & \textbf{12.03} & 42.98 & \textbf{1.04} \\
    \bottomrule
    \end{tabular}
}
\label{results-rlhf}
\vspace{-2mm}
\end{table*}

\subsection{Aligned Model Evaluation Results}

Table~\ref{results-rlhf} summarizes the performance of aligned models based on different reward models. Our $\shortmethodname$ models consistently outperform reward models of similar size, with $\shortmethodname$\textsc{-Qwen3-14B-LoRA-4k} achieving the best overall performance on WildBench, even surpassing the much larger LLaMA-3.3-Nemotron-Super-GenRM (49B) despite having only one-third of the parameters. On MT-Bench, $\shortmethodname$ remains highly competitive, with only a small gap relative to Nemotron. This difference may be partly explained by Nemotron’s training on HelpSteer3, a high-quality dataset featuring multi-turn interactions and multilingual coverage, unlike our $\shortmethodname$-Dataset.

Compared to RM-R1 baselines, our models also deliver stronger results, especially on WildBench. Notably, even $\shortmethodname$\textsc{-Qwen3-4B} outperforms the 14B RM-R1 model on WildBench while remaining competitive on MT-Bench, showcasing the efficiency of our approach. These findings demonstrate that $\shortmethodname$ models consistently provide better performance within their parameter class while scaling effectively to close the gap with much larger models. We also provide additional experiments for on-policy optimization in Section~\ref{sec:on-policy-opt-exps}.

% \subsubsection{Human Evaluation of Reliability of the Reasoning Traces}

% \subsubsection{Human Evaluation of the Summarization of the Reasoning Traces}

\section{Related Work}

\paragraph{Rubric-Based Evaluation Models.}
Recent works leverage explicit rubrics to guide LLM evaluation. \citet{kim2023prometheus} create FeedbackCollection, a fine-grained text evaluation finetuning dataset using detailed rubric for point-wise (direct assessment) evaluation. \cite{kim2024prometheus2} follow-up by adding pair-wise evaluation to the training and found that weight merging performs better than training a jointly trained model. Likewise, LLM-Rubric \citep{hashemi2024llm} prompts an LLM on a human-authored multi-question rubric (e.g., dimensions like naturalness, conciseness, citation quality) and calibrates its outputs via a small model to match human judges. These rubric-driven methods yield fine-grained, interpretable assessments, but their reliance on laboriously constructed rubrics and reference solutions limits scalability and generality \citep{hashemi2024llm,kim2023prometheus}. By contrast, \textsc{R3} eliminates the need for external rubrics, learning reward signals directly in a transparent form to enable broad, rubric-agnostic evaluation.

\paragraph{Preference-Based Reward Models.}
Reward models learned from (implicit or explicit) human preferences—typically via RLHF or related methods—have become a standard alignment approach \citep{rafailov2023direct}. In practice, however, learned RMs often exploit trivial cues: for instance, they tend to favor longer or more elaborate outputs (a well‐known length bias) over brevity \citep{shen2023loose}, and recent analyses show LLM evaluators even “self-recognize” and prefer their own generations over others of equal quality \citep{panickssery2024llm}. \citet{zhu2025charm} further demonstrate “model preference bias” in RMs, whereby certain models’ outputs are systematically overvalued. Such biases and spurious correlations undermine fairness and generalization. New techniques mitigate these issues: DPO recasts RLHF in a simpler optimization framework \citep{rafailov2023direct}, and \citet{vu2024foundational} train FLAMe on 5M+ human judgements across 100+ tasks, achieving stronger OOD generalization and even outperforming GPT-4 on reward-modeling benchmarks. Despite these advances, preference-trained RMs remain large, opaque models tied to specific data, motivating $\shortmethodname$’s interpretable, rubric-free reward formulation as a more transparent alternative.

\paragraph{LLM-as-a-judge Framework.}
Using a pretrained LLM itself as the evaluator has gained popularity due to its zero-shot flexibility \citep{kim2024prometheus2}. However, numerous studies reveal reliability issues. For instance, \citet{wang2024large} find that simply altering the order of candidate responses can drastically flip an LLM judge’s ranking, making one model appear vastly superior or inferior. More broadly, LLM evaluators suffer from hallucinations and entrenched biases; e.g., \citet{panickssery2024llm} show LLM judges systematically score their own outputs higher than others’ (“self-preference” bias), and \citet{zhu2025charm} observe strong model-specific scoring bias in LLM-based evaluation. These flaws undermine trust and consistency in LLM-as-judge systems. $\shortmethodname$ addresses these gaps by providing a fully interpretable reward model that avoids opaque LLM judgments and fixed rubrics, while supporting a broader range of evaluation types for greater flexibility.

\section{Conclusion}
In this paper, we introduce $\shortmethodname$, a novel reward modeling framework that is rubric-agnostic, generalizable across evaluation dimensions, and capable of producing interpretable, reasoning-based score assignments. Leveraging reasoning distillation, targeted dataset curation, and a two-stage quality filtering pipeline, $\shortmethodname$ addresses key limitations of prior reward models in terms of interpretability, controllability, and generalizability. Despite using training datasets that are an order of magnitude smaller than those of many baselines, $\shortmethodname$ models matches or surpasses state-of-the-art performance in reward model benchmarks. Our experiments demonstrate the method’s strong training efficiency and scalability, including effective use of compute-efficient techniques. Policy models that are aligned with our $\shortmethodname$ models also perform better compared to those trained with baseline reward models. By enabling more transparent and adaptable evaluation, $\shortmethodname$ advances robust alignment with diverse human values and real-world applications—paving the way for more trustworthy and versatile reward models.

\bibliography{main}

@inproceedings{kim2023prometheus,
  title={Prometheus: Inducing fine-grained evaluation capability in language models},
  author={Kim, Seungone and Shin, Jamin and Cho, Yejin and Jang, Joel and Longpre, Shayne and Lee, Hwaran and Yun, Sangdoo and Shin, Seongjin and Kim, Sungdong and Thorne, James and others},
  booktitle={The Twelfth International Conference on Learning Representations},
  year={2023}
}

@article{vu2024foundational,
  title={Foundational autoraters: Taming large language models for better automatic evaluation},
  author={Vu, Tu and Krishna, Kalpesh and Alzubi, Salaheddin and Tar, Chris and Faruqui, Manaal and Sung, Yun-Hsuan},
  journal={arXiv preprint arXiv:2407.10817},
  year={2024}
}

@article{lambert2024rewardbench,
  title={Rewardbench: Evaluating reward models for language modeling},
  author={Lambert, Nathan and Pyatkin, Valentina and Morrison, Jacob and Miranda, LJ and Lin, Bill Yuchen and Chandu, Khyathi and Dziri, Nouha and Kumar, Sachin and Zick, Tom and Choi, Yejin and others},
  journal={arXiv preprint arXiv:2403.13787},
  year={2024}
}

@inproceedings{kim2024prometheus2,
  title={Prometheus 2: An Open Source Language Model Specialized in Evaluating Other Language Models},
  author={Kim, Seungone and Suk, Juyoung and Longpre, Shayne and Lin, Bill Yuchen and Shin, Jamin and Welleck, Sean and Neubig, Graham and Lee, Moontae and Lee, Kyungjae and Seo, Minjoon},
  booktitle={Proceedings of the 2024 Conference on Empirical Methods in Natural Language Processing},
  pages={4334--4353},
  year={2024}
}

@article{kim2024biggen,
  title={The biggen bench: A principled benchmark for fine-grained evaluation of language models with language models},
  author={Kim, Seungone and Suk, Juyoung and Cho, Ji Yong and Longpre, Shayne and Kim, Chaeeun and Yoon, Dongkeun and Son, Guijin and Cho, Yejin and Shafayat, Sheikh and Baek, Jinheon and others},
  journal={arXiv preprint arXiv:2406.05761},
  year={2024}
}

@inproceedings{lin2025data,
  title={Data with High and Consistent Preference Difference Are Better for Reward Model},
  author={Lin, Qi and Lu, Hengtong and Yuan, Caixia and Wang, Xiaojie and Jiang, Huixing and Chen, Wei},
  booktitle={Proceedings of the AAAI Conference on Artificial Intelligence},
  volume={39},
  pages={27482--27490},
  year={2025}
}

@inproceedings{hashemi2024llm,
  title={LLM-Rubric: A Multidimensional, Calibrated Approach to Automated Evaluation of Natural Language Texts},
  author={Hashemi, Helia and Eisner, Jason and Rosset, Corby and Van Durme, Benjamin and Kedzie, Chris},
  booktitle={Proceedings of the 62nd Annual Meeting of the Association for Computational Linguistics (Volume 1: Long Papers)},
  pages={13806--13834},
  year={2024}
}

@inproceedings{shen2023loose,
  title={Loose lips sink ships: Mitigating Length Bias in Reinforcement Learning from Human Feedback},
  author={Shen, Wei and Zheng, Rui and Zhan, Wenyu and Zhao, Jun and Dou, Shihan and Gui, Tao and Zhang, Qi and Huang, Xuan-Jing},
  booktitle={Findings of the Association for Computational Linguistics: EMNLP 2023},
  pages={2859--2873},
  year={2023}
}

@article{ankner2024critique,
  title={Critique-out-loud reward models},
  author={Ankner, Zachary and Paul, Mansheej and Cui, Brandon and Chang, Jonathan D and Ammanabrolu, Prithviraj},
  journal={arXiv preprint arXiv:2408.11791},
  year={2024}
}

@article{winata2025preference,
  title={Preference tuning with human feedback on language, speech, and vision tasks: A survey},
  author={Winata, Genta Indra and Zhao, Hanyang and Das, Anirban and Tang, Wenpin and Yao, David D and Zhang, Shi-Xiong and Sahu, Sambit},
  journal={Journal of Artificial Intelligence Research},
  volume={82},
  pages={2595--2661},
  year={2025}
}

@article{zhang2024generative,
  title={Generative verifiers: Reward modeling as next-token prediction},
  author={Zhang, Lunjun and Hosseini, Arian and Bansal, Hritik and Kazemi, Mehran and Kumar, Aviral and Agarwal, Rishabh},
  journal={arXiv preprint arXiv:2408.15240},
  year={2024}
}

@article{pombal2025m,
  title={M-Prometheus: A Suite of Open Multilingual LLM Judges},
  author={Pombal, Jos{\'e} and Yoon, Dongkeun and Fernandes, Patrick and Wu, Ian and Kim, Seungone and Rei, Ricardo and Neubig, Graham and Martins, Andr{\'e} FT},
  journal={arXiv preprint arXiv:2504.04953},
  year={2025}
}

@article{ye2024improving,
  title={Improving reward models with synthetic critiques},
  author={Ye, Zihuiwen and Greenlee-Scott, Fraser and Bartolo, Max and Blunsom, Phil and Campos, Jon Ander and Gall{\'e}, Matthias},
  journal={arXiv preprint arXiv:2405.20850},
  year={2024}
}

@article{zhu2025charm,
  title={CHARM: Calibrating Reward Models With Chatbot Arena Scores},
  author={Zhu, Xiao and Tan, Chenmien and Chen, Pinzhen and Sennrich, Rico and Zhang, Yanlin and Hu, Hanxu},
  journal={arXiv preprint arXiv:2504.10045},
  year={2025}
}

@article{wang2024self,
  title={Self-taught evaluators},
  author={Wang, Tianlu and Kulikov, Ilia and Golovneva, Olga and Yu, Ping and Yuan, Weizhe and Dwivedi-Yu, Jane and Pang, Richard Yuanzhe and Fazel-Zarandi, Maryam and Weston, Jason and Li, Xian},
  journal={arXiv preprint arXiv:2408.02666},
  year={2024}
}

@article{chen2025judgelrm,
  title={Judgelrm: Large reasoning models as a judge},
  author={Chen, Nuo and Hu, Zhiyuan and Zou, Qingyun and Wu, Jiaying and Wang, Qian and Hooi, Bryan and He, Bingsheng},
  journal={arXiv preprint arXiv:2504.00050},
  year={2025}
}

@article{panickssery2024llm,
  title={Llm evaluators recognize and favor their own generations},
  author={Panickssery, Arjun and Bowman, Samuel and Feng, Shi},
  journal={Advances in Neural Information Processing Systems},
  volume={37},
  pages={68772--68802},
  year={2024}
}

@inproceedings{wang2024large,
  title={Large Language Models are not Fair Evaluators},
  author={Wang, Peiyi and Li, Lei and Chen, Liang and Cai, Zefan and Zhu, Dawei and Lin, Binghuai and Cao, Yunbo and Kong, Lingpeng and Liu, Qi and Liu, Tianyu and others},
  booktitle={Proceedings of the 62nd Annual Meeting of the Association for Computational Linguistics (Volume 1: Long Papers)},
  pages={9440--9450},
  year={2024}
}

@inproceedings{wang2024interpretable,
  title={Interpretable Preferences via Multi-Objective Reward Modeling and Mixture-of-Experts},
  author={Wang, Haoxiang and Xiong, Wei and Xie, Tengyang and Zhao, Han and Zhang, Tong},
  booktitle={Findings of the Association for Computational Linguistics: EMNLP 2024},
  pages={10582--10592},
  year={2024}
}

@inproceedings{shridhar2023distilling,
  title={Distilling Reasoning Capabilities into Smaller Language Models},
  author={Shridhar, Kumar and Stolfo, Alessandro and Sachan, Mrinmaya},
  booktitle={Findings of the Association for Computational Linguistics: ACL 2023},
  pages={7059--7073},
  year={2023}
}

@inproceedings{liu2023g,
  title={G-Eval: NLG Evaluation using Gpt-4 with Better Human Alignment},
  author={Liu, Yang and Iter, Dan and Xu, Yichong and Wang, Shuohang and Xu, Ruochen and Zhu, Chenguang},
  booktitle={Proceedings of the 2023 Conference on Empirical Methods in Natural Language Processing},
  pages={2511--2522},
  year={2023}
}

@article{hendrycks2021mmlu,
  title={Measuring Massive Multitask Language Understanding},
  author={Dan Hendrycks and Collin Burns and Steven Basart and Andy Zou and Mantas Mazeika and Dawn Song and Jacob Steinhardt},
  journal={Proceedings of the International Conference on Learning Representations (ICLR)},
  year={2021}
}

@article{suzgun2022bbh,
  title={Challenging BIG-Bench Tasks and Whether Chain-of-Thought Can Solve Them},
  author={Suzgun, Mirac and Scales, Nathan and Sch{\"a}rli, Nathanael and Gehrmann, Sebastian and Tay, Yi and Chung, Hyung Won and Chowdhery, Aakanksha and Le, Quoc V and Chi, Ed H and Zhou, Denny and and Wei, Jason},
  journal={arXiv preprint arXiv:2210.09261},
  year={2022}
}

@article{srivastava2023bigbench,
  title={Beyond the Imitation Game: Quantifying and extrapolating the capabilities of language models},
  author={BIG-bench authors},
  journal={Transactions on Machine Learning Research},
  issn={2835-8856},
  year={2023},
  url={https://openreview.net/forum?id=uyTL5Bvosj},
  note={}
}

@article{chen2025rm,
  title={RM-R1: Reward Modeling as Reasoning},
  author={Chen, Xiusi and Li, Gaotang and Wang, Ziqi and Jin, Bowen and Qian, Cheng and Wang, Yu and Wang, Hongru and Zhang, Yu and Zhang, Denghui and Zhang, Tong and others},
  journal={arXiv preprint arXiv:2505.02387},
  year={2025}
}

@article{guo2025deepseek,
  title={Deepseek-r1: Incentivizing reasoning capability in llms via reinforcement learning},
  author={Guo, Daya and Yang, Dejian and Zhang, Haowei and Song, Junxiao and Zhang, Ruoyu and Xu, Runxin and Zhu, Qihao and Ma, Shirong and Wang, Peiyi and Bi, Xiao and others},
  journal={arXiv preprint arXiv:2501.12948},
  year={2025}
}

@article{muennighoff2025s1,
  title={s1: Simple test-time scaling},
  author={Muennighoff, Niklas and Yang, Zitong and Shi, Weijia and Li, Xiang Lisa and Fei-Fei, Li and Hajishirzi, Hannaneh and Zettlemoyer, Luke and Liang, Percy and Cand{\`e}s, Emmanuel and Hashimoto, Tatsunori},
  journal={arXiv preprint arXiv:2501.19393},
  year={2025}
}

@article{lambert2024tulu3,
  title={T$\backslash$" ulu 3: Pushing frontiers in open language model post-training},
  author={Lambert, Nathan and Morrison, Jacob and Pyatkin, Valentina and Huang, Shengyi and Ivison, Hamish and Brahman, Faeze and Miranda, Lester James V and Liu, Alisa and Dziri, Nouha and Lyu, Shane and others},
  journal={arXiv preprint arXiv:2411.15124},
  year={2024}
}

@article{cui2023ultrafeedback,
  title={Ultrafeedback: Boosting language models with high-quality feedback},
  author={Cui, Ganqu and Yuan, Lifan and Ding, Ning and Yao, Guanming and Zhu, Wei and Ni, Yuan and Xie, Guotong and Liu, Zhiyuan and Sun, Maosong},
  journal={arXiv preprint arXiv:2310.01377},
  year={2023}
}

@article{liu2024skywork,
  title={Skywork-reward: Bag of tricks for reward modeling in llms},
  author={Liu, Chris Yuhao and Zeng, Liang and Liu, Jiacai and Yan, Rui and He, Jujie and Wang, Chaojie and Yan, Shuicheng and Liu, Yang and Zhou, Yahui},
  journal={arXiv preprint arXiv:2410.18451},
  year={2024}
}

@article{lai2024step,
  title={Step-dpo: Step-wise preference optimization for long-chain reasoning of llms},
  author={Lai, Xin and Tian, Zhuotao and Chen, Yukang and Yang, Senqiao and Peng, Xiangru and Jia, Jiaya},
  journal={arXiv preprint arXiv:2406.18629},
  year={2024}
}

@article{zeng2025acecoder,
  title={ACECODER: Acing Coder RL via Automated Test-Case Synthesis},
  author={Zeng, Huaye and Jiang, Dongfu and Wang, Haozhe and Nie, Ping and Chen, Xiaotong and Chen, Wenhu},
  journal={arXiv preprint arXiv:2502.01718},
  year={2025}
}

@article{wang2018glue,
  title={GLUE: A multi-task benchmark and analysis platform for natural language understanding},
  author={Wang, Alex and Singh, Amanpreet and Michael, Julian and Hill, Felix and Levy, Omer and Bowman, Samuel R},
  journal={arXiv preprint arXiv:1804.07461},
  year={2018}
}

@article{wang2019superglue,
  title={Superglue: A stickier benchmark for general-purpose language understanding systems},
  author={Wang, Alex and Pruksachatkun, Yada and Nangia, Nikita and Singh, Amanpreet and Michael, Julian and Hill, Felix and Levy, Omer and Bowman, Samuel},
  journal={Advances in neural information processing systems},
  volume={32},
  year={2019}
}

@article{fabbri2021summeval,
  title={Summeval: Re-evaluating summarization evaluation},
  author={Fabbri, Alexander R and Kry{\'s}ci{\'n}ski, Wojciech and McCann, Bryan and Xiong, Caiming and Socher, Richard and Radev, Dragomir},
  journal={Transactions of the Association for Computational Linguistics},
  volume={9},
  pages={391--409},
  year={2021},
  publisher={MIT Press One Rogers Street, Cambridge, MA 02142-1209, USA journals-info~…}
}

@article{wang2023evaluating,
  title={Evaluating open-qa evaluation},
  author={Wang, Cunxiang and Cheng, Sirui and Guo, Qipeng and Yue, Yuanhao and Ding, Bowen and Xu, Zhikun and Wang, Yidong and Hu, Xiangkun and Zhang, Zheng and Zhang, Yue},
  journal={Advances in Neural Information Processing Systems},
  volume={36},
  pages={77013--77042},
  year={2023}
}

@article{achiam2023gpt,
  title={Gpt-4 technical report},
  author={Achiam, Josh and Adler, Steven and Agarwal, Sandhini and Ahmad, Lama and Akkaya, Ilge and Aleman, Florencia Leoni and Almeida, Diogo and Altenschmidt, Janko and Altman, Sam and Anadkat, Shyamal and others},
  journal={arXiv preprint arXiv:2303.08774},
  year={2023}
}

@article{li2023towards,
  title={Towards general text embeddings with multi-stage contrastive learning},
  author={Li, Zehan and Zhang, Xin and Zhang, Yanzhao and Long, Dingkun and Xie, Pengjun and Zhang, Meishan},
  journal={arXiv preprint arXiv:2308.03281},
  year={2023}
}

@inproceedings{shahapure2020cluster,
  title={Cluster quality analysis using silhouette score},
  author={Shahapure, Ketan Rajshekhar and Nicholas, Charles},
  booktitle={2020 IEEE 7th international conference on data science and advanced analytics (DSAA)},
  pages={747--748},
  year={2020},
  organization={IEEE}
}

@article{yang2025qwen3,
  title={Qwen3 Technical Report},
  author={Yang, An and Li, Anfeng and Yang, Baosong and Zhang, Beichen and Hui, Binyuan and Zheng, Bo and Yu, Bowen and Gao, Chang and Huang, Chengen and Lv, Chenxu and others},
  journal={arXiv preprint arXiv:2505.09388},
  year={2025}
}

@article{liu2024rm,
  title={RM-bench: Benchmarking reward models of language models with subtlety and style},
  author={Liu, Yantao and Yao, Zijun and Min, Rui and Cao, Yixin and Hou, Lei and Li, Juanzi},
  journal={arXiv preprint arXiv:2410.16184},
  year={2024}
}

@article{narayan2018don,
  title={Don't give me the details, just the summary! topic-aware convolutional neural networks for extreme summarization},
  author={Narayan, Shashi and Cohen, Shay B and Lapata, Mirella},
  journal={arXiv preprint arXiv:1808.08745},
  year={2018}
}

@article{zhang2024benchmarking,
  title={Benchmarking large language models for news summarization},
  author={Zhang, Tianyi and Ladhak, Faisal and Durmus, Esin and Liang, Percy and McKeown, Kathleen and Hashimoto, Tatsunori B},
  journal={Transactions of the Association for Computational Linguistics},
  volume={12},
  pages={39--57},
  year={2024},
  publisher={MIT Press One Broadway, 12th Floor, Cambridge, Massachusetts 02142, USA~…}
}

@article{abdin2025phi,
  title={Phi-4-reasoning Technical Report},
  author={Abdin, Marah and Agarwal, Sahaj and Awadallah, Ahmed and Balachandran, Vidhisha and Behl, Harkirat and Chen, Lingjiao and de Rosa, Gustavo and Gunasekar, Suriya and Javaheripi, Mojan and Joshi, Neel and others},
  journal={arXiv preprint arXiv:2504.21318},
  year={2025}
}

@article{sen1968estimates,
  title={Estimates of the regression coefficient based on Kendall's tau},
  author={Sen, Pranab Kumar},
  journal={Journal of the American statistical association},
  volume={63},
  number={324},
  pages={1379--1389},
  year={1968},
  publisher={Taylor \& Francis}
}

@inproceedings{wang2025helpsteerpreference,
  title={HelpSteer2-Preference: Complementing Ratings with Preferences},
  author={Zhilin Wang and Alexander Bukharin and Olivier Delalleau and Daniel Egert and Gerald Shen and Jiaqi Zeng and Oleksii Kuchaiev and Yi Dong},
  booktitle={The Thirteenth International Conference on Learning Representations},
  year={2025},
  url={https://openreview.net/forum?id=MnfHxPP5gs}
}

@article{winata2024metametrics,
  title={Metametrics: Calibrating metrics for generation tasks using human preferences},
  author={Winata, Genta Indra and Anugraha, David and Susanto, Lucky and Kuwanto, Garry and Wijaya, Derry Tanti},
  journal={arXiv preprint arXiv:2410.02381},
  year={2024}
}

@article{zhang2024general,
  title={General Preference Modeling with Preference Representations for Aligning Language Models},
  author={Zhang, Yifan and Zhang, Ge and Wu, Yue and Xu, Kangping and Gu, Quanquan},
  journal={arXiv preprint arXiv:2410.02197},
  year={2024}
}

@misc{skyworkcritic2024,
  title={Skywork Critic Model Series},
  author={Shiwen, Tu and Liang, Zhao and Liu, Chris Yuhao and Zeng, Liang and Liu, Yang},
  year={2024},
  month={September},
  howpublished={\url{https://huggingface.co/Skywork}},
  url={https://huggingface.co/Skywork},
}

@article{yu2025improve,
  title={Improve LLM-as-a-Judge Ability as a General Ability},
  author={Yu, Jiachen and Sun, Shaoning and Hu, Xiaohui and Yan, Jiaxu and Yu, Kaidong and Li, Xuelong},
  journal={arXiv preprint arXiv:2502.11689},
  year={2025}
}

@article{li2019acute,
  title={Acute-eval: Improved dialogue evaluation with optimized questions and multi-turn comparisons},
  author={Li, Margaret and Weston, Jason and Roller, Stephen},
  journal={arXiv preprint arXiv:1909.03087},
  year={2019}
}

@article{stureborg2024large,
  title={Large language models are inconsistent and biased evaluators},
  author={Stureborg, Rickard and Alikaniotis, Dimitris and Suhara, Yoshi},
  journal={arXiv preprint arXiv:2405.01724},
  year={2024}
}

@article{hu2022lora,
  title={Lora: Low-rank adaptation of large language models.},
  author={Hu, Edward J and Shen, Yelong and Wallis, Phillip and Allen-Zhu, Zeyuan and Li, Yuanzhi and Wang, Shean and Wang, Lu and Chen, Weizhu and others},
  journal={ICLR},
  volume={1},
  number={2},
  pages={3},
  year={2022}
}

@inproceedings{zheng2024llamafactory,
  title={LlamaFactory: Unified Efficient Fine-Tuning of 100+ Language Models},
  author={Yaowei Zheng and Richong Zhang and Junhao Zhang and Yanhan Ye and Zheyan Luo and Zhangchi Feng and Yongqiang Ma},
  booktitle={Proceedings of the 62nd Annual Meeting of the Association for Computational Linguistics (Volume 3: System Demonstrations)},
  address={Bangkok, Thailand},
  publisher={Association for Computational Linguistics},
  year={2024},
  url={http://arxiv.org/abs/2403.13372}
}

@inproceedings{kwon2023efficient,
  title={Efficient Memory Management for Large Language Model Serving with PagedAttention},
  author={Woosuk Kwon and Zhuohan Li and Siyuan Zhuang and Ying Sheng and Lianmin Zheng and Cody Hao Yu and Joseph E. Gonzalez and Hao Zhang and Ion Stoica},
  booktitle={Proceedings of the ACM SIGOPS 29th Symposium on Operating Systems Principles},
  year={2023}
}

@article{jia2023zero,
  title={Zero-shot faithfulness evaluation for text summarization with foundation language model},
  author={Jia, Qi and Ren, Siyu and Liu, Yizhu and Zhu, Kenny Q},
  journal={arXiv preprint arXiv:2310.11648},
  year={2023}
}

@inproceedings{zhong2022towards,
  title={Towards a Unified Multi-Dimensional Evaluator for Text Generation},
  author={Zhong, Ming and Liu, Yang and Yin, Da and Mao, Yuning and Jiao, Yizhu and Liu, Pengfei and Zhu, Chenguang and Ji, Heng and Han, Jiawei},
  booktitle={Proceedings of the 2022 Conference on Empirical Methods in Natural Language Processing},
  pages={2023--2038},
  year={2022}
}

@article{team2024qwen2,
  title={Qwen2 technical report},
  author={Team, Qwen},
  journal={arXiv preprint arXiv:2407.10671},
  year={2024}
}

@article{rafailov2023direct,
  title={Direct preference optimization: Your language model is secretly a reward model},
  author={Rafailov, Rafael and Sharma, Archit and Mitchell, Eric and Manning, Christopher D and Ermon, Stefano and Finn, Chelsea},
  journal={Advances in neural information processing systems},
  volume={36},
  pages={53728--53741},
  year={2023}
}

@article{wang2025helpsteer3,
  title={HelpSteer3-Preference: Open Human-Annotated Preference Data across Diverse Tasks and Languages},
  author={Wang, Zhilin and Zeng, Jiaqi and Delalleau, Olivier and Shin, Hoo-Chang and Soares, Felipe and Bukharin, Alexander and Evans, Ellie and Dong, Yi and Kuchaiev, Oleksii},
  journal={arXiv preprint arXiv:2505.11475},
  year={2025}
}

@article{zheng2023judging,
  title={Judging llm-as-a-judge with mt-bench and chatbot arena},
  author={Zheng, Lianmin and Chiang, Wei-Lin and Sheng, Ying and Zhuang, Siyuan and Wu, Zhanghao and Zhuang, Yonghao and Lin, Zi and Li, Zhuohan and Li, Dacheng and Xing, Eric and others},
  journal={Advances in neural information processing systems},
  volume={36},
  pages={46595--46623},
  year={2023}
}

@article{lin2024wildbench,
  title={Wildbench: Benchmarking llms with challenging tasks from real users in the wild},
  author={Lin, Bill Yuchen and Deng, Yuntian and Chandu, Khyathi and Brahman, Faeze and Ravichander, Abhilasha and Pyatkin, Valentina and Dziri, Nouha and Bras, Ronan Le and Choi, Yejin},
  journal={arXiv preprint arXiv:2406.04770},
  year={2024}
}

@article{dubois2024length,
  title={Length-Controlled AlpacaEval: A Simple Way to Debias Automatic Evaluators},
  author={Dubois, Yann and Galambosi, Bal{\'a}zs and Liang, Percy and Hashimoto, Tatsunori B},
  journal={arXiv preprint arXiv:2404.04475},
  year={2024}
}
\bibliographystyle{tmlr}

\newpage

\appendix

\section{Limitations}
\label{sec:limit}
In our experiments, we limit our exploration to models with up to 14B parameters due to resource constraints. We also include smaller models in our study, aiming to shed light on scaling behavior and its impact on performance. Larger models, such as those with 32B parameters or more, are left for future investigation.

% \section{Evaluation Datasets}
% \label{sec:3.5}

\section{Details about Datasets}
\label{apx:datasets}

\subsection{Details About Reward Model Training Datasets}
\label{apx:detail-rm-train-datasets}

\subsubsection{Details About Dataset Sampling}
\label{axp:detail-dataset-sampling}

The following describes a more detailed process of the three-stage sampling process to extract the most diverse examples:

\begin{enumerate}
    \item \textbf{Embedding and Preprocessing}. We begin by embedding each instance using a semantic representation that combines its task instruction and input text to capture the sample's semantics across topics. Specifically, we use the \texttt{gte-Qwen2-7B-instruct} model~\citep{li2023towards} to compute embeddings over $h(x^{(j)}) = t^{(j)}\oplus i^{(j)}$, where $\oplus$ denotes string concatenation. The resulting embedding $Emb(h(x^{(j)})) = q^{(j)}$ is used to measure similarity and diversity in semantic space during clustering.
    
    \item \textbf{Cluster Determination and Assignment}. To identify an appropriate number of groups $k^* \in \{k_{\min}, \dots, k_{\max}\}$, we select the value of \(k\) that maximizes the average Silhouette score~\citep{shahapure2020cluster}. Here we choose $k_{min} = 3$ and $k_{max} = 10$. If the dataset includes labeled subcategories (e.g., topics or task types), clustering is applied independently within each subcategory to preserve intra-category diversity. The Silhouette score for a sample $x^{(j)}$ is defined as $s_j = \frac{v_j - w_j}{\max(v_j, w_j)}$ where $v_j$ is the mean distance between $x_j$ and other points in the same cluster, and $w_j$ is the mean distance to the nearest cluster not containing $x^{(j)}$. We select the optimal number of clusters $k^*$ by
    \begin{align}
    k^* = \arg\max_{k \in \{k_{\min}, \dots, k_{\max}\}} \frac{1}{|\mathcal{D}|} \sum_{j=1}^{|\mathcal{D}|} s_j^{(k)},
    \end{align}
    where $s_j^{(k)}$ is the Silhouette score of sample $x^{(j)}$ under the clustering configuration with $k$ clusters.
    
    \item \textbf{Stratified Sampling with Maximal Marginal Relevance (MMR)}. We perform stratified sampling from each cluster with a minimum of 10 samples per cluster. For each cluster $C$ with centroid $q_{C}$:
    \begin{itemize}
        \item We retain the first 25\% of samples based on the closest to the cluster centroid, to ensure topical relevance, i.e., $R_{closest} = Top_{\lfloor{0.25 \cdot |C|\rfloor}} \{x \in C\text{ } | \text{ }\norm{Emb(x) - q_C}_{2}\}$;
        \item The next 75\% of the samples are selected via MMR, which balances relevance and diversity among the already selected samples. Let $R$ denote the set of already selected examples, in which initially $R = R_{closest}$. To sample the next candidate $x \in C \setminus R$, we compute the MMR score as:
        \begin{align}
        \text{MMR}(x) = \lambda \cdot \text{sim}(x, q_{C}) - (1 - \lambda) \cdot \max_{x_r \in R} \text{sim}(x, x_r),
        \end{align}
        where $\text{sim}(\cdot, \cdot)$ denotes cosine similarity, and $\lambda \in [0, 1]$ is a tunable trade-off parameter, in which we set $\lambda = 0.5$ to balance relevance and diversity. The next selected example is $x^* = \arg\max_{x \in C \setminus R} \text{MMR}(x)$.
    \end{itemize}
\end{enumerate}
5
\subsubsection{Dataset Size and Composition}

\begin{table*}[!ht]
    \centering
    \caption{Dataset size and composition of the top 7 source datasets at each stage of filtering. FC = Feedback Collection, PC = Preference Collection.}
    \label{tab:dataset-size}
    \resizebox{\textwidth}{!}{
    \begin{tabular}{lr|cccccccccc}\toprule
    & \textbf{Count} & \textbf{Tulu3} & \textbf{AceCodePair} & \textbf{Math-step-DPO} & \textbf{FC} &\textbf{PC} & \textbf{UltraFeedback} & \textbf{Skywork} \\\midrule
    $\mathcal{D}_{20k}$ & 20,000 & 0.18 & 0.15 & 0.15 &0.13 & 0.10 & 0.10 & 0.10 \\
    $\mathcal{D}_{14k}$ (Filter Step 1) &13,772 &0.19 &0.20 &0.21 &0.09 &0.07 &0.06 &0.11 \\
    $\mathcal{D}_{4k}$ (Filter Step 2) &3,949 &0.13 &0.28 &0.19 &0.12 &0.03 &0.12 &0.05 \\
    \bottomrule
    \end{tabular}
    }
\end{table*}

Table~\ref{tab:dataset-size} showcases the composition of $\mathcal{D}_{20k}$, $\mathcal{D}_{14k}$, and $\mathcal{D}_{4k}$.

\subsubsection{Prompt and Response Length}\label{apx:prompt-response-len}

\begin{table*}[!ht]
    \centering
    \caption{Length (white-space separated word count) distribution of our dataset. Response length includes DeepSeek-R1 thinking tokens along with the short response, which contains an explanation and the score assigned.}
    \label{tab:prompt-response-len}
    \resizebox{0.6\textwidth}{!}{
    \begin{tabular}{lrrr}
    \toprule
    &\textbf{Prompt Length} &\textbf{Response Length} \\\midrule
    $\mathcal{D}_{20k}$ & 504 $\pm$ 302 &850 $\pm$ 847 \\
    $\mathcal{D}_{14k}$ (Filter Step 1) & 497 $\pm$ 413 &729 $\pm$ 538 \\
    $\mathcal{D}_{4k}$ (Filter Step 2) &442 $\pm$ 224 &851 $\pm$ 599 \\
    \bottomrule
    \end{tabular}
    }
\end{table*}

In Table~\ref{tab:prompt-response-len} we document the length distribution of our dataset.

\subsubsection{Label Distribution}

In Table~\ref{tab:label-distribution} we show the label distribution of our dataset across different filtering stages. Our raw dataset has balanced distribution within each evaluation type. In $\mathcal{D}_{14k}$ (Filter Step 1), binary labels are slightly biased towards ``false'' and pair-wise labels are slightly biased towards ``Response 1''. In $\mathcal{D}_{4k}$ (Filter Step 2), binary labels are slightly biased towards ``true'' and and pair-wise labels are slightly biased towards ``Response 1''. Point-wise scores are also biased towards middle values (i.e., ``3'').
\label{tab:label-distribution}
\begin{table*}[!ht]
    \centering
    \caption{Dataset label statistics distribution across the filtering process.}
    \small
    \begin{tabular}{lrr|rr|rrrrrr}\toprule
    &\multicolumn{2}{c}{\textbf{Binary}} &\multicolumn{2}{c}{\textbf{Pair-wise}} &\multicolumn{5}{c}{\textbf{Point-wise}} \\\cmidrule{2-10}
    &true &false &resp. 1 &resp. 2 & 1 & 2 & 3 & 4 & 5 \\\midrule
    $\mathcal{D}_{20k}$ &0.024 &0.026 &0.340 &0.335 &0.047 &0.053 &0.055 &0.058 &0.062 \\
    $\mathcal{D}_{14k}$ (Filter Step 1) &0.024 &0.031 &0.429 &0.354 &0.040 &0.036 &0.021 &0.038 &0.027 \\
    $\mathcal{D}_{4k}$ (Filter Step 2) &0.033 &0.022 &0.365 &0.304 &0.035 &0.048 &0.700 &0.061 &0.046 \\
    \bottomrule
    \end{tabular}
\end{table*}

\subsection{Details About Reward Model Evaluation Datasets}
\label{apx:detail-rm-eval-datasets}

RM-Bench~\citep{liu2024rm} contains 1.3K instances across four domains—Chat, Safety, Math, and Code—each with prompts at three difficulty levels. RewardBench~\citep{lambert2024rewardbench} includes 3K preference pairs in four categories: Chat, Chat-Hard, Safety, and Reasoning, where we report both category-wise and overall accuracy. For summarization, we use a human-annotated subset of XSUM~\citep{narayan2018don} with labels on faithfulness (binary), coherence, and relevance (Likert 1–5) following \citet{zhang2024benchmarking}; we compute accuracy for faithfulness and Kendall-Tau~\citep{sen1968estimates} correlations for coherence and relevance. FeedbackBench~\citep{kim2023prometheus}, the test split of FeedbackCollection, contains 1K rubrics, 200 instructions, and 1K responses, evaluated via Kendall-Tau correlation. For STEM reasoning, we construct MMLU-STEM Binary~\citep{hendrycks2021mmlu} from a subset.\footnote{\url{https://huggingface.co/datasets/TIGER-Lab/MMLU-STEM}}
\paragraph{RM-Bench}~\citep{liu2024rm} is a reward model evaluation benchmark consisting of 1.3K instances that cover four domains: Chat, Safety, Math, and Code. Each instance consists of three prompts categorized by difficulty level: easy, medium, and hard. We measure the accuracy on each domain and difficulty level, along with the overall average accuracy. 

\paragraph{RewardBench.}\citep{lambert2024rewardbench}  is a popular reward model evaluation benchmark consists of 3K instances of preference pairs on four categories: Chat, Chat-Hard, Safety, Reasoning. We measure the accuracy on each category along with the overall average accuracy.

\paragraph{XSUM.}~\citep{narayan2018don}  is a news summarization dataset. For our evaluation, we use a subset that has been annotated by human evaluators across three criteria: faithfulness (binary), coherence (Likert scale 1–5), and relevance (Likert scale 1–5), following the annotation protocol of \citet{zhang2024benchmarking}. We measure the Kendall-Tau~\citep{sen1968estimates} correlation for coherence and relevance, while we measure accuracy for faithfulness.

\paragraph{FeedbackBench.}~\citep{kim2023prometheus}  is the test split of FeedbackCollection introduced with the Prometheus model for evaluating point-wise tasks. It contains 1K score rubrics, 200 instructions, and 1K responses that do not overlap with the train data. We measure the Kendall-Tau~\citep{sen1968estimates} correlation as previously done by ~\citet{kim2023prometheus}. 

\paragraph{MMLU-STEM Binary.}~\citep{hendrycks2021mmlu} is a STEM-subject related subset\footnote{\url{https://huggingface.co/datasets/TIGER-Lab/MMLU-STEM}.} of the MMLU benchmark with multiple-choice questions from various branches of knowledge. Given four potential choices and one correct answer, we convert it to a binary evaluation task. For each original question, we evaluate model's response given the correct and separate a randomly selected incorrect answer. We measure the overall accuracy, along with the accuracy on each subject. Unless otherwise specified, all references to MMLU-STEM in this work refer to our MMLU-STEM Binary benchmark.

\paragraph{BBH Binary.}~\citep{suzgun2022bbh}  is a collection of 27 non-trivial reasoning-like tasks sourced from BigBench~\citep{srivastava2023bigbench} with a total of 6.7K instances. The format of the tasks can be multiple choice or short string completion. Similar to MMLU-STEM, we include a copy of the data with the correct response and a copy with the incorrect response. Details of the dataset generation process is in Appendix~\ref{apx:evaluation-dataset-generation}. We measure the overall accuracy. Unless otherwise specified, all references to BBH in this work refer to our BBH Binary benchmark.

\subsection{Details About Policy Alignment Evaluation Datasets}
\label{apx:detail-align-eval-datasets}

MT Bench contains 80 samples from 8 diverse categories (Writing, Roleplay, Extraction, STEM, Humanities, Reasoning, Math and Coding), each with two turns. WildBench, on the other han,d contains 1024 diverse real-world variable-turn prompts relating to Creative, Planning/Reasoning, Data Analysis/Math, Information/Advice seeking and Coding/Debugging.

\section{Prompt Template}\label{apx:prompt-templates}

\subsection{Rubric Generation Template}\label{apx:rubric-gen-template}
For point-wise tasks, we generate rubric with Likert score from 1 to 5 using the following template.
\begin{tcolorbox}[colback=gray!10,colframe=black,title=Rubric generation template,breakable]
You are an expert evaluator. Given a defined task, analyze the task and create a rubric using a Likert scale from 1 to 5 to that will help to perform the given task. \\
Please follow these steps: \\
1. Explain the criteria for distinguishing between the scores (e.g., how a score of 1 differs from a score of 5). \\
2. Based on your analysis, generate a rubric in JSON format with the Likert scale ranging from 1 to 5, including descriptions for each score. \\
3. Ensure that the rubric is clear, actionable, and covers key aspects of the task. \\
\\
\#\#\# TASK\\
\{task\_instruction\}\\
\\
\#\#\# INPUT\\
\{input/question\}\\
\\
\#\#\# EXAMPLE RUBRICS (Unrelated Tasks)\\
\{sample\_rubrics\}\\

\#\#\# RUBRIC FOR CURRENT TASK\\
\end{tcolorbox}

\subsection{Point-wise Evaluation}\label{apx:point-wise-template}
For point-wise tasks where the judge model needs to assign a score for a response from 1-5, we use the following template.
\vspace{-0.5em}
\begin{tcolorbox}[colback=gray!10,colframe=black,title=Pointwise evaluation prompt template,breakable]
Evaluate the response based on the given task, input, response, and evaluation rubric.\\
Provide a fair and detailed assessment following the rubric.\\
\\
\#\#\# TASK\\
\{task\_instruction\}\\
\\
\#\#\# INPUT\\
\{input/question\}\\
\\
\#\#\# RESPONSE\\
\{response\}\\

\#\#\# EVALUATION RUBRIC\\
1: \{score\_of\_1\_description\} \\
2: \{score\_of\_2\_description\} \\
3: \{score\_of\_3\_description\} \\
4: \{score\_of\_4\_description\} \\
5: \{score\_of\_5\_description\} \\
\\
\#\#\# OUTPUT FORMAT\\
Return a JSON response in the following format:  \\
\\
\{ \\
"explanation": "Explanation of why the response received a particular score",\\ 
"score": "Score assigned to the response based on the rubric between 1 to 5"\\
\}\\
\\
\#\#\# EVALUATION\\
\end{tcolorbox}

\subsection{Pair-wise Evaluation}\label{apx:pair-wise-template}
For pair-wise tasks where the judge model needs to compare against two responses, we use the following template.
\vspace{-0.5em}
\begin{tcolorbox}[colback=gray!10,colframe=black,title=Pairwise evaluation prompt template,breakable]
Evaluate the response based on the given task, input, two responses, and evaluation rubric.\\
Provide a fair and detailed assessment following the rubric.  \\
\\
\#\#\# TASK \\
\{task\_instruction\}  \\
\\
\#\#\# INPUT \\
\{input/question\}\\
\\
\#\#\# RESPONSE 1 \\
\{response\_1\}\\
\\
\#\#\# RESPONSE 2 \\
\{response\_2\}\\
\\
\#\#\# EVALUATION RUBRIC \\
Response 1: Response 1 provided better response, rejecting Response 2. \\
Response 2: Response 2 provided better response, rejecting Response 1.  \\
\\
\#\#\# OUTPUT FORMAT \\
Return a JSON response in the following format:  \\
\\
\{\\
"explanation": "Explanation of why one response is preferred over the other", \\
"score": "Final selection between 'Response 1' or 'Response 2'" \\
\}  \\
\\
\#\#\# EVALUATION
\end{tcolorbox}

For rubrics, we include three variations and uniformly randomly sample from them when creating our dataset.
\vspace{-0.5em}
\begin{tcolorbox}[colback=gray!10,colframe=black,title=Pairwise evaluation rubric variation 1,breakable]
\{\\
    "response\_1": "Response 1 is the preferred choice over Response 2.",\\
    "response\_2": "Response 2 is the preferred choice over Response 1."\\
\}\\
\end{tcolorbox}

\begin{tcolorbox}[colback=gray!10,colframe=black,title=Pairwise evaluation rubric variation 2,breakable]
\{\\
    "response\_1": "Response 1 provided better response, rejecting Response 2.",\\
    "response\_2": "Response 2 provided better response, rejecting Response 1."\\
\}\\
\end{tcolorbox}

\begin{tcolorbox}[colback=gray!10,colframe=black,title=Pairwise evaluation rubric variation 3,breakable]
\{\\
    "response\_1": "Response 1 is superior, meaning Response 2 is not chosen.",\\
    "response\_2": "Response 2 is superior, meaning Response 1 is not chosen."\\
\}\\
\end{tcolorbox}

\subsection{Binary Evaluation}\label{apx:binary-template}
For binary tasks where the judge model needs to classify true or false to the response, we use the following template.
\vspace{-0.5em}
\begin{tcolorbox}[colback=gray!10,colframe=black,title=Binary evaluation prompt template,breakable]
Evaluate the response based on the given task, input, response, and evaluation rubric.  \\
Provide a fair and detailed assessment following the rubric.  \\
\\
\#\#\# TASK \\
\{task\_instruction\} \\
\\
\#\#\# INPUT \\
\{input/question\} \\ 
\\
\#\#\# RESPONSE \\
\{response\}  \\
\\
\#\#\# EVALUATION RUBRIC \\
true: The response accurately reflects the correct answer based on the input.\\
false: The response does not accurately reflect the correct answer based on the input.  \\
\\
\#\#\# OUTPUT FORMAT \\
Return a JSON response in the following format:  \\
\\
\{ \\
"explanation": "Explanation of why the answer is true or false", \\
"score": "Final boolean answer between true or false" \\
\}\\
\\
\#\#\# EVALUATION
\end{tcolorbox}

For rubrics, we include three variations and uniformly randomly sample from them when creating our dataset.
\vspace{-0.5em}
\begin{tcolorbox}[colback=gray!10,colframe=black,title=Binary evaluation rubric variation 1,breakable]
\{\\
    "true": "The answer correctly corresponds to the given input.",\\
    "false": "The answer does not correctly correspond to the given input."\\
\}\\
\end{tcolorbox}

\begin{tcolorbox}[colback=gray!10,colframe=black,title=Binary evaluation rubric variation 2,breakable]
\{\\
    "true": "The response accurately reflects the correct answer based on the input.",\\
    "false": "The response does not accurately reflect the correct answer based on the input."\\
\}\\
\end{tcolorbox}

\begin{tcolorbox}[colback=gray!10,colframe=black,title=Binary evaluation rubric variation 3,breakable]
\{\\
    "true": "The answer is appropriate and correctly derived from the input.",\\
    "false": "The answer is inappropriate or incorrectly derived from the input."\\
\}\\
\end{tcolorbox}

\section{Example Prompts and Responses}

\subsection{Point-wise Evaluation}
\vspace{-0.5em}
\begin{tcolorbox}[colback=gray!10,colframe=black,title=Prompt example for pointwise evaluation (sourced from FeedbackCollection),breakable]
Evaluate the response based on the given task, input, response, and evaluation rubric.\\
Provide a fair and detailed assessment following the rubric.\\
\\
\#\#\# TASK\\
Does the final conclusion drawn by the response hold up to logical scrutiny and provide a correct solution for an instruction with a definite answer?\\
\\
\#\#\# INPUT\\
Imagine a situation where there is a debate going on regarding the total number of prime numbers between 1 and 100. Your task is to determine the correct count and also provide the list. The debate is currently revolving around three different answers - 25, 26 and 27 prime numbers.\\
\\
\#\#\# RESPONSE\\
The total number of prime numbers between 1 and 100 is definitely 30. Prime numbers are those numbers that only have 1 and the number itself as factors. We don't need to list down each of them because it's already known that there are 30 prime numbers between 1 and 100. It's simple mathematics, and there is no need for any debate on this matter. The discussion around the number being 25, 26, or 27 is irrelevant and baseless. So, the conclusion is there are 30 prime numbers between 1 and 100.\\

\#\#\# EVALUATION RUBRIC\\
1: The concluding answer from the model is entirely false and devoid of logical foundation.\\
2: The concluding answer from the model has major flaws that seriously compromise its validity. \\
3: The concluding answer from the model bears considerable mistakes that demand significant rectification. \\
4: The concluding answer from the model has slight inaccuracies, but these are simple to fix and do not greatly affect its overall validity. \\
5: The concluding answer from the model is wholly correct and logically sound.\\
\\
\#\#\# OUTPUT FORMAT\\
Return a JSON response in the following format:  \\
\\
\{ \\
"explanation": "Explanation of why the response received a particular score",\\ "score": "Score assigned to the response based on the rubric between 1 to 5"\\
\}\\
\\
\#\#\# EVALUATION\\
\end{tcolorbox}

\begin{tcolorbox}[colback=gray!10,colframe=black,title=Example response from R1 (summarized) used for SFT,breakable]
<think>\\
Okay, let me evaluate this response. The user asked about the number of primes between 1 and 100, with the debate being 25, 26, or 27. The response says there are 30 primes, which is definitely wrong. The actual count is 25. The response claims it's "simple mathematics" and dismisses the debate, but that's incorrect. The answer provided is entirely false, and the reasoning is flawed because they didn't list or verify the primes. The rubric says a score of 1 is for entirely false and no logical foundation. This fits because the answer is wrong and the logic is absent. So, score 1.\\
</think>\\
\{\\
"explanation": "The response concludes there are 30 prime numbers between 1 and 100, which is factually incorrect. The actual number is 25. The claim that 'it’s already known there are 30 primes' is baseless, and the dismissal of the debate over 25, 26, or 27 as 'irrelevant' ignores the correct answer (25). The response fails to provide logical reasoning, a correct list, or acknowledgment of established mathematical facts. This demonstrates a complete lack of validity and logical foundation.", \\
"score": "1"\\
\}
\end{tcolorbox}

\subsection{Pair-wise Evaluation}
\vspace{-0.5em}
% \begin{CJK}{UTF8}{gbsn}  % for Chinese characters with pdfLatex
\begin{tcolorbox}[colback=gray!10,colframe=black,title=Prompt example for pairwise evaluation (sourced from Tulu-3-sft),breakable]
Evaluate the response based on the given task, input, two responses, and evaluation rubric.\\
Provide a fair and detailed assessment following the rubric.  \\
\\
\#\#\# TASK \\
Evaluate the factual accuracy of the response. Consider whether the information provided is correct, up-to-date, and free from errors or misconceptions.  \\
\\
\#\#\# INPUT \\
Come up with a question and reasoning that would justify this answer: Raphael Sbarge Your ENTIRE response should be in Chinese, no other language is allowed. Come up with a question and reasoning that would justify this answer: Raphael Sbarge  \\
\\
\#\#\# RESPONSE 1 \\\begin{CJK}{UTF8}{gbsn}
Question: 《Blade》这部吸血鬼题材的电影中，扮演主角V的演员是谁？  \\
\\
Reasoning: 电影《Blade》是一部著名的吸血鬼题材作品，Raphael Sbarge在这部电影中饰演了主角V，这是一个非常有影响力的角色。因此，这个问题和回答完美契合，Raphael Sbarge正是该角色的扮演者。  \\
\\
\#\#\# RESPONSE 2 \\
问题：谁是演 'Once Upon a Time' 中的 "金Pinocchio/Archie Hopper" 一角的演员？  \\
\\
理由：Raphael Sbarge 是美国演员，他最著名的角色之一就是美国电视剧 'Once Upon a Time' 中的金Pinocchio/Archie Hopper。   \\\end{CJK}
\\
\#\#\# EVALUATION RUBRIC \\
Response 1: Response 1 provided better response, rejecting Response 2. \\
Response 2: Response 2 provided better response, rejecting Response 1.  \\
\\
\#\#\# OUTPUT FORMAT \\
Return a JSON response in the following format:  \\
\\
\{\\
"explanation": "Explanation of why one response is preferred over the other", \\
"score": "Final selection between 'Response 1' or 'Response 2'" \\
\}  \\
\\
\#\#\# EVALUATION
\end{tcolorbox}

\begin{CJK}{UTF8}{gbsn}
\begin{tcolorbox}[colback=gray!10,colframe=black,title=Pairwise example response from R1 (summarized) used for SFT,breakable]
<think>\\
好的，我现在需要评估两个回答，看看哪个在事实准确性上更好。任务是根据提供的输入，判断哪个回答更准确、最新且没有错误。输入是让用户想出一个问题和理由来证明答案是Raphael Sbarge。  \\

首先看Response 1的问题：“《Blade》这部吸血鬼题材的电影中，扮演主角V的演员是谁？”理由里说Raphael Sbarge在这部电影中饰演了主角V。但根据我的了解，《Blade》系列电影的主角Blade是由Wesley Snipes扮演的，而“V”可能是指另一部电影《V字仇杀队》里的主角V，但那个是由Hugo Weaving饰演的。Raphael Sbarge确实参演过《Blade》吗？我需要确认一下。查资料发现，Raphael Sbarge确实在1998年的《刀锋战士》（Blade）中出演过一个角色，但并不是主角V，而是次要角色Dr. Curtis Webb。所以Response 1的问题和理由存在错误，把主角V的扮演者错误地归为Sbarge，这显然是错误的。  \\
\\
接下来看Response 2的问题：“谁是演 'Once Upon a Time' 中的 '金Pinocchio/Archie Hopper' 一角的演员？”理由指出Sbarge在该剧中扮演这个角色。根据资料，Raphael Sbarge确实在《童话镇》（Once Upon a Time）中扮演了Archie Hopper，也就是木偶匹诺曹的人类形态。这个信息是正确的，所以Response 2的问题和理由都是事实正确的。  \\
\\
根据评分标准，应该选择事实准确的回答。显然Response 2正确，而Response 1有错误，因此Response 2更优。\\
</think>```json \\
\{ \\
"explanation": "Response 1存在事实错误：《Blade》系列电影的主角V并非由Raphael Sbarge饰演（实际由Wesley Snipes饰演），且该电影中Raphael Sbarge仅饰演次要角色Dr. Curtis Webb。Response 2则正确指出Raphael Sbarge在《Once Upon a Time》中饰演Archie Hopper/金Pinocchio，这一信息符合事实且无错误。因此，Response 2在事实准确性上明显优于Response 1。", \\
"score": "Response 2" \\
\} ```
\end{tcolorbox}
\end{CJK}
\subsection{Binary Evaluation}
\vspace{-0.5em}
\begin{tcolorbox}[colback=gray!10,colframe=black,title=Prompt example for binary evaluation (sourced from evouna),breakable]
Evaluate the response based on the given task, input, response, and evaluation rubric.  \\
Provide a fair and detailed assessment following the rubric.  \\
\\
\#\#\# TASK \\
Your task is to determine whether the given answer response is correct based on the query input.  \\
\\
\#\#\# INPUT \\
who was the killer in the movie i know what you did last summer  \\
\\
\#\#\# RESPONSE \\
Ben Willis  \\
\\
\#\#\# EVALUATION RUBRIC \\
true: The response accurately reflects the correct answer based on the input.\\
false: The response does not accurately reflect the correct answer based on the input.  \\
\\
\#\#\# OUTPUT FORMAT \\
Return a JSON response in the following format:  \\
\\
\{ \\
"explanation": "Explanation of why the answer is true or false", \\
"score": "Final boolean answer between true or false" \\
\}\\
\\
\#\#\# EVALUATION
\end{tcolorbox}

\begin{tcolorbox}[colback=gray!10,colframe=black,title=Prompt example for binary evaluation (sourced from evouna),breakable]
<think>\\
Okay, let's tackle this evaluation. The user is asking whether the killer in the movie "I Know What You Did Last Summer" is Ben Willis, as stated in the response. First, I need to recall the plot of the movie to verify this.  From what I remember, the movie revolves around a group of friends who hit someone with their car and then cover it up. The killer is later revealed to be seeking revenge for that incident. The main antagonist is indeed Ben Willis, who is also known as the Fisherman. He's the father of the person they hit, and he faked his own death to frame someone else. So, the response "Ben Willis" is correct. The answer should be true.\\
</think>\\
\{\\
"explanation": "In the movie 'I Know What You Did Last Summer,' the killer is revealed to be Ben Willis, also known as the Fisherman. He is the father of the victim the group accidentally hit with their car and covers up. Ben Willis fakes his death and seeks revenge, making the response accurate.", \\
"score": "true" \\
\}
\end{tcolorbox}

\section{Explanation Trace Summarization Details}\label{apx:trace-summarization}
First, we perform inference using the model to obtain the initial reasoning trace. This training trace is then passed through the model once more; conditioned on the prompt shown in ``Prompt for Summarization Tracing'' to generate a concise version. The second inference produces a shortened reasoning trace by removing redundant or unnecessary reasoning steps while preserving the original tone, style, and logical progression.
\begin{tcolorbox}[colback=gray!10,colframe=black,title=Prompt for Reasoning Trace Summarization,breakable]
Shorten the following reasoning trace by removing redundant or unnecessary thinking loops while preserving the exact same tone, style, and progression of thought. Output only the shortened reasoning trace without any explanation. \\

\{DeepSeek-R1 Reasoning Trace\}\\
\end{tcolorbox}

\section{Human Evaluation of Reasoning Traces}\label{sec:human-eval-trace}
We recruit five annotators to annotate approximately 2\% of $\mathcal{D}_{4k}$, which was stratified sampled from various dataset sources, to verify both the reliability of the reasoning traces and the quality of the trace summarization. Details of the annotations setup, metrics we use to annotate, the experiments, and results are in Appendix~\ref{apx:human-annotation-details}. We find on average the reasoning traces score $2.9\pm0.2$ (out of 3, higher better) in factual correctness, $2.8\pm0.2$ in logical coherence (n=93). The faithfulness of the summary scores averages $2.8\pm0.5$ and the style consistency scores $2.7\pm0.4$ (n=84). These results confirm the high quality reasoning traces used in our dataset.

\section{Training Hyper-parameters}
\label{sec:hyperparam}

For all of our experiments, we use 4 A800 80GB GPUs.

We use LLaMA-Factory~\citep{zheng2024llamafactory} to perform SFT for all $\shortmethodname{}$ models. We set the maximum sequence length to 8192, with a learning rate of $1\mathrm{e}{-5}$, trained for 5 epochs using a cosine learning rate scheduler. The batch size per device is 16. For $\shortmethodname{}$ LoRA models, we use LoRA rank of 64 and alpha of 128. For inference, we use vLLM~\citep{kwon2023efficient} using the recommended inference configuration from Qwen3 and Phi-4-reasoning-plus. 

\section{Evaluation Prompt}\label{apx:evaluation-dataset-generation}

Since RewardBench and FeedbackBench are of pair-wise and point-wise evaluation format, they do not require extra processing to format into our prompt template. For both MMLU-STEM and BBH, since we are converting them to binary evaluation, we need to sample negative responses to augment the dataset.

The original MMLU-STEM consists of multiple-choice questions. We simply randomly sample a wrong answer as the negative. For subtasks of BBH that are also in the format of multiple-choice questions, we do the same.

There are four tasks that require custom adaptation for negative label sampling:

\paragraph{DyckLanguages} is a task where models are tasked to complete un-closed parentheses of different types. To sample negatives, with equal chance, we randomly delete, swap, or insert a symbol that appears in the context.

\paragraph{WordSorting} is a task where models are tasked to sort a set of unordered words. We randomly swap a pair of words from the target order to create the negative.

\paragraph{MultistepArithmeticTwo} is a task where models are expected to perform arithmetic calculations involving 8 single-digit operands. We calculate the mean and standard deviation of the label distribution, and randomly sample a number within the distribution.

\paragraph{ObjectCounting} is a task where models are expected to count the number of objects (possibly a subset of all mentioned objects) mentioned in a sentence. We calculate the mean and standard deviation of the label count distribution, and randomly sample a number within the distribution.

\section{Human Annotation Details}\label{apx:human-annotation-details}
We stratified-sample 100 instances of data, and have the authors of the paper annotate the quality of the reasoning and reasoning summarizations. In total we have 5 annotators, annotating a total of around 2\% of $\mathcal{D}_{4k}$.

\subsection{Reliability of Reasoning Trace}
To ensure reasoning trace is reliable, we define two metrics \textbf{Factual Correctness} and \textbf{Logical Coherence} to ensure consistent labeling:
\paragraph{Factual Correctness} (Scale: 1–3) assesses whether the statements in the reasoning trace are true and supported by external knowledge or evidence. When scoring, treat retrievable evidence or commonsense facts as acceptable grounding.
\begin{enumerate}
    \item (Incorrect)    Contains one or more clear factual errors or hallucinations that undermine the trace. May lead to incorrect conclusions or mislead the model.
    \item (Partially Correct)    Most statements are accurate, but minor factual errors or unverifiable claims exist. Does not change the final conclusion, but may reduce trace reliability.
    \item (Fully Correct)    All statements are factually accurate and supported by known facts, context, or ground truth. No hallucinations or inaccuracies.
\end{enumerate}

\paragraph{Logical Coherence} measures whether the reasoning steps logically follow from each other and form a coherent argument or thought process. Judge based on internal consistency, not factuality. A trace can be factually wrong but still logically coherent.

\begin{enumerate}
    \item (Incoherent)    Trace is illogical, disjointed, or internally inconsistent. Steps may contradict, skip crucial logic, or appear arbitrary.
    \item (Somewhat Coherent)    Mostly logical, but has minor gaps, unclear transitions, or weak justifications. Still understandable, but less robust as supervision.
    \item (Fully Coherent)    All steps follow logically and consistently. No missing steps, contradictions, or unjustified jumps in reasoning. A smooth, interpretable chain.
\end{enumerate}

In Table~\ref{tab:human-annotation-reasoning} we show detailed annotation results across annotators.

\begin{table*}[!ht]
    \centering
    \caption{Human annotation results on reasoning trace \textbf{factual correctness} and \textbf{logical coherence} (out of 3, higher better).}
\begin{tabular}{lrrrrrr}\toprule
&\textbf{Annotator 1 }&\textbf{Annotator 2} &\textbf{Annotator 3} &\textbf{Annotator 4} &\textbf{Average} \\\midrule
Factual Correctness &3 $\pm$ 0.2 &3 $\pm$ 0 &2.9 $\pm$ 0.3 &2.8 $\pm$ 0.5 &2.9 $\pm$ 0.2 \\
Logical Coherence &2.9 $\pm$ 0.4 &2.6 $\pm$ 0.7 &2.9 $\pm$ 0.3 &2.7 $\pm$ 0.5 &2.8 $\pm$ 0.2 \\
Count &27 &10 &28 &28 &- \\
\bottomrule
\end{tabular}
    \label{tab:human-annotation-reasoning}
\end{table*}

\subsection{Reasoning Trace Summary Quality}
During dataset curation, we use GPT-4.1 mini to summarize the reasoning traces that are too long. We want to measure \textbf{faithfulness} and \textbf{style similarity}.

\paragraph{Faithfulness} measures how well the summary covers the ideas of the original reasoning trace

\begin{enumerate}
    \item (Unfaithful) Omits key reasoning or introduces incorrect logic. Could mislead a model or change the original meaning.
    \item (Partially Faithful)    Minor omissions or slightly altered emphasis, but preserves the general logic and outcome. Acceptable for training.
    \item (Fully Faithful)    Captures all core and necessary reasoning steps accurately. No hallucinations, distortions, or omissions of crucial logic.
\end{enumerate}

\paragraph{Style Similarity} includes similar tone, level of formality, structured markers (``first'', ``therefore''), or domain-specific phrasing.

\begin{enumerate}
    \item (Completely different)   Omits all tone, level of formality, etc. from original trace
    \item (Somewhat similar style)    Somewhat similar in terms of tone, level of formality, etc. from original trace
    \item (Same style)    Same style with the original reasoning trace
\end{enumerate}

In Table~\ref{tab:human-annotation-reasoning-summary} we show detailed annotation results across annotators.

\begin{table*}[!ht]
    \centering
    \caption{Human annotation results on reasoning trace summary \textbf{faithfulness} and \textbf{style similarity} (out of 3, higher better).}
    \small
\begin{tabular}{lrrrrrrr}\toprule
 &\textbf{Annotator 1} &\textbf{Annotator 2} &\textbf{Annotator 3} &\textbf{Annotator 4} &\textbf{Annotator 5} &\textbf{Average} \\\midrule
Faithfulness &2.6 $\pm$ 0.8 &2.8 $\pm$ 0.5 &2.8 $\pm$ 0.4 &2.7 $\pm$ 0.8 &3.0 $\pm$ 0.0 &2.8 $\pm$ 0.5 \\
Style similarity &2.5 $\pm$ 0.6 &3.0 $\pm$ 0.2 &2.7 $\pm$ 0.5 &2.8 $\pm$ 0.4 &2.7 $\pm$ 0.5 &2.7 $\pm$ 0.4 \\
Count &20 &26 &25 &6 &7 & - \\
\bottomrule
\end{tabular}
    \label{tab:human-annotation-reasoning-summary}
\end{table*}

\section{Detailed Results Breakdowns}
\label{apx:detailed-results-break}

\subsection{RM-Bench \& Reward Bench}
Additional results presented in Table~\ref{tab:results-rmbench-supplementary} and Table~\ref{tab:results-rewardbench-supplementary} are derived from the findings reported in \cite{chen2025rm}.

\subsection{BBH Binary \& MMLU-STEM Binary}
Table~\ref{tab:bbh-supplementary} reports additional results from our BBH Binary and MMLU-STEM Binary.

\subsection{XSUM and FeedbackBench}
Table~\ref{tab:XSUM-supplementary} reports additional results, reproduced from prior work including \citet{jia2023zero}, to provide broader context and facilitate direct comparison across XSUM and FeedbackBench benchmarks.

\begin{table*}[!ht]
\centering
\caption{Comparison of existing models with $\shortmethodname$ on RM-Bench. \textbf{Bolded numbers} indicate the best-performing results within each group section independently.}
\resizebox{0.6\textwidth}{!}{
    \begin{tabular}{lcccc|ccc|c}
    \toprule
    \textbf{Model} & \multicolumn{4}{c|}{Domain} & \multicolumn{3}{c|}{Difficulty} & Overall \\
    & Chat & Math & Code & Safety & Easy & Medium & Hard & Avg. \\ 
    \midrule
    % ArmoRM & ? & ? & ? & ? & ? & ? & ? & ?\\
    Base Models \\ \midrule
    Qwen2.5-7B & 59.9 & 52.6 & 50.9 & 72.4 & 66.4 & 60.3 & 50.1 & 58.9 \\
    DeepSeek-Distilled-Qwen-14B & 65.0 & 85.4	& 69.0 & 85.1 & 84.3	& 78.7	& 65.4 & 76.1 \\
    Qwen3-4B & 64.3	& 84.0	& 62.0	& 85.6	& 82.6	& 74.5	& 64.8	& 74.0 \\
    Qwen3-8B&  63.2	& 80.5	& 61.0	& 84.8	& 80.1	& 73.4	& 63.5	& 72.4 \\
    Qwen3-14B & 76.4 & 88.0	& 70.7 & 91.0 & 87.9 & 84.3 & 72.3 & 81.5 \\
    Phi-4-R$^+$-14B & 73.4 & 87.3 & 70.7 &	85.7 & 86.5 & 80.2 & 71.1 &	79.3 \\
    \midrule
    Scalar RMs \\ \midrule
    steerlm-70b & 56.4 & 53.0 & 49.3 & 51.2 & 48.3 & 54.9 & 54.3 & 52.5 \\
    tulu-v2.5-70b-preference-mix-rm & 58.2 & 51.4 & 55.5 & 87.1 & 72.8 & 65.6 & 50.7 & 63.0 \\
    Mistral-7B-instruct-Unified-Feedback & 56.5 & 58.0 & 51.7 & 86.8 & 87.1 & 67.3 & 35.3 & 63.2 \\
    RM-Mistral-7B & 57.4 & 57.0 & 52.7 & 87.2 & 88.6 & 67.1 & 34.9 & 63.5 \\
    Eurus-RM-7b & 59.9 & 60.2 & 56.9 & 86.5 & 87.2 & 70.2 & 40.2 & 65.9 \\
    internlm2-7b-reward & 61.7 & \textbf{71.4} & 49.7 & 85.5 & 85.4 & 70.7 & 45.1 & 67.1 \\
    Skywork-Reward-Gemma-2-27B & 69.5 & 54.7 & 53.2 & 91.9 & 78.0 & 69.2 & 54.9 & 67.3 \\
    ArmoRM-Llama3-8B-v0.1 & 67.8 & 57.5 & 53.1 & 92.4 & 82.2 & 71.0 & 49.8 & 67.7 \\
    GRM-llama3-8B-sftreg & 62.7 & 62.5 & 57.8 & 90.0 & 83.5 & 72.7 & 48.6 & 68.2 \\
    internlm2-20b-reward & 63.1 & 66.8 & 56.7 & 86.5 & 82.6 & 71.6 & 50.7 & 68.3 \\
    Llama-3-OffsetBias-RM-8B & \textbf{71.3} & 61.9 & 53.2 & 89.6 & 84.6 & 72.2 & 50.2 & 69.0 \\
    Nemotron-340B-Reward & 71.2 & 59.8 & \textbf{59.4} & 87.5 & 81.0 & 71.4 & \textbf{56.1} & 69.5 \\
    URM-Llama-3.1-8B & 71.2 & 61.8 & 54.1 & 93.1 & 84.0 & 73.2 & 53.0 & 70.0 \\
    Skywork-Reward-Llama-3.1-8B & 69.5 & 60.6 & 54.5 & \textbf{95.7} & 89.0 & 74.7 & 46.6 & 70.1 \\
    infly/INF-ORM-Llama3.1-70B & 66.3 & 65.6 & 56.8 & 94.8 & \textbf{91.8} & \textbf{76.1} & 44.8 & \textbf{70.9} \\ \midrule
    Generative RMs \\ \midrule
    tulu-v2.5-dpo-13b-chatbot-arena-2023 & 64.9 & 52.3 & 50.5 & 62.3 & 82.8 & 60.2 & 29.5 & 57.5 \\
    tulu-v2.5-dpo-13b-nectar-60k & 56.3 & 52.4 & 52.6 & 73.8 & 86.7 & 64.3 & 25.4 & 58.8 \\
    stablelm-2-12b-chat & 67.2 & 54.9 & 51.6 & 65.2 & 69.1 & 63.5 & 46.6 & 59.7 \\
    tulu-v2.5-dpo-13b-stackexchange-60k & 66.4 & 49.9 & 54.2 & 69.0 & 79.5 & 63.0 & 37.2 & 59.9 \\
    Nous-Hermes-2-Mistral-7B-DPO & 58.8 & 55.6 & 51.3 & 73.9 & 69.5 & 61.1 & 49.1 & 59.9 \\
    Claude-3-5-sonnet-20240620 & 62.5 & 62.6 & 54.5 & 64.4 & 73.8 & 63.4 & 45.9 & 61.0 \\
    tulu-v2.5-dpo-13b-hh-rlhf-60k & 68.4 & 51.1 & 52.3 & 76.5 & 53.6 & 63.0 & 69.6 & 62.1 \\
    tulu-2-dpo-13b & 66.4 & 51.4 & 51.8 & 85.4 & 86.9 & 66.7 & 37.7 & 63.8 \\
    SOLAR-10.7B-Instruct-v1.0 & 78.6 & 52.3 & 49.6 & 78.9 & 57.5 & 67.6 & 69.4 & 64.8 \\
    Llama3.1-70B-Instruct & 64.3 & 67.3 & 47.5 & 83.0 & 74.7 & 67.8 & 54.1 & 65.5 \\
    Skywork-Critic-Llama-3.1-70B & 71.4 & 64.6 & 56.8 & 94.8 & 85.6 & 73.7 & 56.5 & 71.9 \\
    GPT-4o-0806 & 67.2 & 67.5 & 63.6 & 91.7 & 83.4 & 75.6 & 58.7 & 72.5 \\
    Gemini-1.5-pro & 71.6 & 73.9 & 63.7 & 91.3 & 83.1 & 77.6 & 64.7 & 75.2 \\ 
    Prometheus-7B-v2.0 & 46.0 & 52.6 & 47.6 & 73.9 & 68.8 & 54.9 & 41.3 & 55.0 \\ 
    JudgeLRM & 59.9 & 59.9 & 51.9 & 87.3 & 73.2 & 76.6 & 54.8 & 64.7 \\
    RM-R1-Qwen-Instruct-7B & 66.6 & 67.0 & 54.6 & 92.6 & 79.2 & 71.7 & 59.7 & 70.2 \\
    RM-R1-DeepSeek-Distilled-Qwen-7B & 64.0 & 83.9 & 56.2 & 85.3 & 75.9 & 73.1 & 68.1 & 72.4 \\
    RM-R1-Qwen-Instruct-14B & \textbf{75.6} & 75.4 & 60.6 & 93.6 & 82.6 & 77.5 & 68.8 & 76.1 \\
    RM-R1-Qwen-Instruct-32B & 75.3 & 80.2 & 66.8 & 93.9 & 86.3 & 80.5 & 70.4 & 79.1 \\
    RM-R1-DeepSeek-Distilled-Qwen-14B & 71.8 & 90.5 & 69.5 & 94.1 & 86.2 & 83.6 & 74.4 & 81.5 \\
    RM-R1-DeepSeek-Distilled-Qwen-32B & 74.2 & \textbf{91.8} & \textbf{74.1} & \textbf{95.4} & \textbf{89.5} & \textbf{85.4} & \textbf{76.7} & \textbf{83.9} \\
    \midrule
    $\shortmethodname$ Models (Ours) \\ \midrule
    $\shortmethodname\textsc{-Qwen3-4B-LoRA-4k}$ &  68.2 & 93.4 & 72.6 & 85.4 & 87.4 & 81.3 & 71.1 & 79.9\\
    $\shortmethodname\textsc{-Qwen3-4B-LoRA-14k}$ &  66.9 & 92.2 & 72.7 & 86.5 & 86.9 & 81.5 & 70.3 & 79.6 \\
    $\shortmethodname\textsc{-Qwen3-4B-4k}$ & 68.9 & 92.3 & 72.5 & 86.5 & 86.5 & 81.4 & 72.3 & 80.0 \\
    $\shortmethodname\textsc{-Qwen3-4B-14k}$ & 67.9 & 93.0 & 74.7 & 86.9 & 88.8 & 81.9 & 71.1 & 80.6 \\
    $\shortmethodname\textsc{-Qwen3-8B-LoRA-4k}$ & 68.9 & 93.5 & 75.2 & 88.1 & 88.2 & 83.8 & 72.4 & 81.4 \\
    $\shortmethodname\textsc{-Qwen3-8B-LoRA-14k}$ & 68.9 & 92.9 & 75.0 & 88.9 & 89.0 & 83.2 & 72.1 & 81.4  \\
    % $\shortmethodname$\textsc{-Qwen3-8B-LoRA-14k}$ & 70.9 & 90.1 & 71.8 & 89.1 & 88.9 & 82.5 & 70.0 & 80.5 \\
    $\shortmethodname\textsc{-Qwen3-8B-4k}$ & 70.8 & 92.9 & 74.2 & 89.2 & 87.9 & 83.4 & 74.0 & 81.8 \\
    $\shortmethodname\textsc{-Qwen3-8B-14k}$ & 69.1 & 93.2 & 75.9 & 87.6 & 89.0 & 83.4 & 71.9 & 81.4 \\
    $\shortmethodname\textsc{-Qwen3-14B-LoRA-4k}$ & 74.6 & 93.9 & 78.7 & \textbf{89.8} & 90.2 & 86.3 & \textbf{76.2} & \textbf{84.2} \\
    $\shortmethodname\textsc{-Qwen3-14B-LoRA-14k}$ & 73.8 & 93.6 & 77.4 & 89.0 & 89.7 & 85.9 & 74.8 & 83.5 \\
    $\shortmethodname\textsc{-Qwen3-14B-4k}$ & 74.0 & 93.7 & 77.2 & 89.3 & 89.7 & 85.3 & 75.6 & 83.6 \\
    $\shortmethodname$\textsc{-Qwen3-14B-14k} & 73.4 & 93.8 & \textbf{79.1} & 89.5 & \textbf{90.3} & \textbf{86.6} & 74.9 & 84.0 \\ 
    $\shortmethodname$\textsc{-Phi-4-R$^+$-14B-LoRA-4k} & 71.4 & \textbf{94.4} & 78.2 & 86.2 & 88.7 & 84.3 & 74.7 & 82.5 \\ 
    $\shortmethodname$\textsc{-Phi-4-R$^+$-14B-LoRA-14k} & 73.2 & 90.9 & 73.7 & 85.3 & 87.7 & 82.9 & 71.7 & 80.8 \\ 
    $\shortmethodname$\textsc{-Phi-4-R$^+$-14B-4k} & \textbf{74.9} & 90.7 & 74.1 & 86.6 & 87.9 & 83.3 & 73.5 & 81.6  \\ 
    $\shortmethodname$\textsc{-Phi-4-R$^+$-14B-14k} & 74.5 & 93.0 & 77.5 & 84.8 & 89.3 & 84.7 & 73.3 & 82.5 \\
    $\shortmethodname\textsc{-Qwen2.5-7B-LoRA-4K}$ & 59.6 & 60.2 & 49.4 & 76.3 & 71.2 & 63.1 & 49.8 & 61.4 \\
    % $\shortmethodname\textsc{-Qwen2.5-7B-LoRA-14K}$ &  \\
    $\shortmethodname\textsc{-Qwen2.5-7B-4K}$ & 69.6 & 75.5 & 59.8 & 86.9 & 80.2 & 74.2 & 64.5 & 73.0 \\
    $\shortmethodname\textsc{-Qwen2.5-7B-14K}$ & 66.8 & 82.0 & 65.0 & 87.0 & 83.8 & 76.8 & 64.9 & 75.2 \\
    $\shortmethodname\textsc{-DeepSeek-Distilled-Qwen-14B-LoRA-4K}$ & 69.0 & 90.3 & 70.5 & 85.8 & 85.9 & 81.6 & 69.3 & 78.9 \\
    $\shortmethodname\textsc{-DeepSeek-Distilled-Qwen-14B-LoRA-14K}$ & 68.0 & 90.8 & 71.2 & 86.7 & 87.0 & 81.8 & 68.9 & 79.2 \\
    $\shortmethodname\textsc{-DeepSeek-Distilled-Qwen-14B-4K}$ & 73.0 & 92.2 & 77.1 & 86.3 & 88.5 & 84.1 & 73.9 & 82.1 \\
    $\shortmethodname\textsc{-DeepSeek-Distilled-Qwen-14B-14K}$ & 71.7 & 93.0 & 78.4 & 86.4 & 89.3 & 84.7 & 73.1 & 82.4 \\
    \midrule
    Proprietary Models \\ \midrule
    GPT-4.1 mini & 67.6 & 73.0 & 71.3 & 90.7 & 87.0 & 78.4 & 61.7 & 75.7 \\
    GPT-o4 mini & 77.6 & \textbf{93.0} & 80.8 & \textbf{93.4} & \textbf{92.0} & \textbf{88.7} & \textbf{78.0} & \textbf{86.2} \\
    GPT-5 mini & \textbf{88.0} & 92.9 & \textbf{91.1} & 78.0 & 77.4 & 85.8 & 96.4 & 92.4 \\
    DeepSeek-R1 & 78.6 & 66.2 & 81.9 & 88.7 & 86.9 & 82.2 & 67.3 & 78.8 \\
    \bottomrule
    \end{tabular}
}
\label{tab:results-rmbench-supplementary}
\end{table*}

\begin{table*}[!ht]
\centering
\caption{Comparison of existing models with $\shortmethodname$ on RewardBench using pair-wise scoring. \textbf{Bolded numbers} indicate the best-performing results within each group section independently.}
\resizebox{0.6\textwidth}{!}{
    \begin{tabular}{lcccc|c}
    \toprule
    \textbf{Models} & Chat & Chat Hard & Safety & Reasoning & Avg. \\ 
    \midrule
    Base Models \\ \midrule
    Qwen2.5-7B & 86.3 & 60.1 & 77.8 & 72.1 & 74.1 \\
    DeepSeek-Distilled-Qwen-14B & 91.1 & 72.8 & 83.0 & 88.5 & 83.8 \\
    Qwen3-4B & 88.6 & 65.1 & 83.8 &	89.7 & 81.8 \\
    Qwen3-8B& 89.1 & 70.0 & 82.7 & 88.9 & 82.7 \\
    Qwen3-14B & 93.6 & 78.5 & 87.7 & 95.1 & 88.7 \\
    Phi-4-R$^+$-14B & 92.7 & 79.6 & 85.85 & 90.1 & 87.1 \\
    \midrule
    Scalar RMs \\ \midrule
    Eurus-RM-7b & 98.0 & 65.6 & 81.4 & 86.3 & 82.8 \\
    Internlm2-7b-reward & \textbf{99.2} & 69.5 & 87.2 & 94.5 & 87.6 \\
    SteerLM-RM 70B & 91.3 & 80.3 & 92.8 & 90.6 & 88.8 \\
    Cohere-0514 & 96.4 & 71.3 & 92.3 & 97.7 & 89.4 \\
    Internlm2-20b-reward & 98.9 & 76.5 & 89.5 & 95.8 & 90.2 \\
    ArmoRM-Llama3-8B-v0.1 & 96.9 & 76.8 & 90.5 & 97.3 & 90.4 \\
    Nemotrom-4-340B-Reward & 95.8 & 87.1 & 91.5 & 93.6 & 92.0 \\
    Skywork-Reward-Llama-3.1-8B & 95.8 & 87.3 & 90.8 & 96.2 & 92.5 \\
    Skywork-Reward-Gemma-2-27B & 95.8 & \textbf{91.4} & 91.9 & 96.1 & 93.8 \\
    infly/INF-ORM-Llama3.1-70B & 96.6 & 91.0 & \textbf{93.6} & \textbf{99.1} & \textbf{95.1} \\ \midrule
    Generative RMs \\ \midrule
    Llama3.1-8B-Instruct & 85.5 & 48.5 & 75.6 & 72.1 & 70.4 \\
    Llama3.1-70B-Intruct & \textbf{97.2} & 70.2 & 82.8 & 86.0 & 84.0 \\
    Llama3.1-405B-Intruct & \textbf{97.2} & 74.6 & 77.6 & 87.1 & 84.1 \\
    Claude-3-5-sonnet-20240620 & 96.4 & 74.0 & 81.6 & 84.7 & 84.2 \\
    GPT-4o-0806 & 96.1 & 76.1 & 86.6 & 88.1 & 86.7 \\
    Gemini-1.5-pro & 92.3 & 80.6 & 87.9 & 92.0 & 88.2 \\
    Self-taught-evaluator-llama3.1-70B & 96.9 & 85.1 & 89.6 & 88.4 & 90.0 \\
    SFR-LLaMa-3.1-70B-Judge-r & 96.9 & 84.8 & 91.6 & \textbf{97.6} & 92.7 \\
    Skywork-Critic-Llama-3.1-70B & 96.6 & \textbf{87.9} & \textbf{93.1} & 95.5 & \textbf{93.3} \\
    Prometheus-7B-v2.0 & 90.2 & 45.6 & 75.8 & 74.6 & 71.6\\ 
    m-Prometheus-14B & 93.6 & 59.0 & 85.1 & 84.8 & 80.6 \\ 
    JudgeLRM & 92.9 & 56.4 & 78.2 & 73.6 & 75.2 \\
    SynRM & 38.0 & 82.5 & 74.1 & 87.1 & 70.4\\
    % \midrule
    RM-R1-DeepSeek-Distilled-Qwen-7B & 88.9 & 66.2 & 78.4 & 87.0 & 80.1 \\
    RM-R1-Qwen-Instruct-7B & 94.1 & 74.6 & 85.2 & 86.7 & 85.2 \\
    RM-R1-Qwen-Instruct-14B & 93.6 & 80.5 & 86.9 & 92.0 & 88.2 \\
    RM-R1-DeepSeek-Distilled-Qwen-14B & 91.3 & 79.4 & 89.3 & 95.5 & 88.9 \\
    RM-R1-DeepSeek-Distilled-Qwen-32B & 95.3 & 80.3 & 91.1 & 96.8 & 90.9 \\
    RM-R1-Qwen-Instruct-32B & 95.3 & 83.1 & 91.9 & 95.2 & 91.4 \\
    \midrule
    $\shortmethodname$ Models (Ours) \\ \midrule
    $\shortmethodname\textsc{-Qwen3-4B-LoRA-4K}$ & 91.1 & 74.4 & 85.6 & 95.5 & 86.7 \\
    $\shortmethodname\textsc{-Qwen3-4B-LoRA-14K}$ & 90.4 & 75.2 & 85.7 & 96.1 & 86.9 \\
    $\shortmethodname\textsc{-Qwen3-4B-4K}$ & 88.3 & 77.4 & 86.1 & 95.3 & 86.8 \\
    $\shortmethodname\textsc{-Qwen3-4B-14K}$ & 92.4 & 76.0 & 85.8 & 95.7 & 87.5 \\
    $\shortmethodname\textsc{-Qwen3-8B-LoRA-4K}$ & 93.2 & 76.6 & 87.0 & 96.3 & 88.3 \\
    $\shortmethodname\textsc{-Qwen3-8B-LoRA-14K}$ & 93.0 & 76.2 & 87.6 & 96.4 & 88.3 \\
    $\shortmethodname\textsc{-Qwen3-8B-4K}$ & 91.6 & 79.8 & 87.7 & 95.8 & 88.7 \\
    $\shortmethodname\textsc{-Qwen3-8B-14K}$ & 93.8 & 78.6 & 86.3 & 96.7 & 88.8 \\
    $\shortmethodname\textsc{-Qwen3-14B-LoRA-4K}$ & 93.6 & \textbf{85.1} & \textbf{88.7} & 96.8 & \textbf{91.0} \\
    $\shortmethodname\textsc{-Qwen3-14B-LoRA-14K}$ & 92.9 & 82.8 & 88.2 & \textbf{96.9} & 90.2 \\
    $\shortmethodname\textsc{-Qwen3-14B-4K}$ & 92.6 & 81.0 & 88.4 & 96.6 & 89.7 \\
    $\shortmethodname\textsc{-Qwen3-14B-14K}$ & 93.3 & 79.7 & 88.4 & \textbf{96.9} & 89.6 \\
    $\shortmethodname\textsc{-Phi-4-R$^+$-14B-LoRA-4K}$ & 90.6 & 76.5 & 86.8 & 96.5 & 87.6 \\
    $\shortmethodname\textsc{-Phi-4-R$^+$-14B-LoRA-14K}$ & 93.4 & 79.1 & 85.2 & 94.3 & 88.0 \\
    $\shortmethodname\textsc{-Phi-4-R$^+$-14B-4K}$ & 92.6 & 79.0 & 85.8 & 96.3 & 88.4 \\
    $\shortmethodname\textsc{-Phi-4-R$^+$-14B-14K}$ & \textbf{94.5} & 78.0 & 86.6 & 96.5 & 88.9 \\
    $\shortmethodname\textsc{-Qwen2.5-7B-LoRA-4K}$ & 83.1 & 67.0 & 79.4 & 73.2 & 75.7 \\
    $\shortmethodname\textsc{-Qwen2.5-7B-4K}$ & 85.9 & 75.3 & 85.5 & 85.1 & 82.9 \\
    $\shortmethodname\textsc{-Qwen2.5-7B-14K}$ & 91.4 & 73.8 & 85.1 & 90.6 & 85.2 \\
    $\shortmethodname\textsc{-DeepSeek-Distilled-Qwen-14B-LoRA-4K}$ & 90.8 & 75.6 & 84.6 & 93.1 & 86.0 \\
    $\shortmethodname\textsc{-DeepSeek-Distilled-Qwen-14B-LoRA-14K}$ & 92.4 & 75.2 & 84.7 & 93.8 & 86.5 \\
    $\shortmethodname\textsc{-DeepSeek-Distilled-Qwen-14B-4K}$ & 89.7 & 78.7 & 86.0 & 95.5 & 87.5 \\
    $\shortmethodname\textsc{-DeepSeek-Distilled-Qwen-14B-14K}$ & 92.3 & 77.8 & 86.8 & 95.6 & 88.1 \\
    \midrule
    Propretiary Models \\ \midrule
    GPT-4.1 mini & \textbf{96.1} & 75.2 & 87.0 & 89.6 & 87.0 \\ 
    GPT-o4 mini & 95.3 & \textbf{81.8} & 91.6 & \textbf{98.4} & \textbf{91.8} \\
    GPT-5 mini & 95.3 & 81.6 & \textbf{92.0} & \textbf{98.4} & \textbf{91.8} \\
    DeepSeek-R1 & 93.6 & 79.2 & 86.9 & 97.4 & 89.3 \\
    \bottomrule
    \end{tabular}
    }
\label{tab:results-rewardbench-supplementary}
\end{table*}

\begin{table*}[!ht]
\centering
\caption{Comparison of existing models with $\shortmethodname{}$ on BBH \& MMLU-STEM binary. \textbf{Bolded numbers} indicate the best-performing results within each group section independently.}
\label{tab:bbh-supplementary}
% \resizebox{1\textwidth}{!}{
    \begin{tabular}{lcc}
    \toprule
    \textbf{Models} & \textbf{BBH Binary} & \textbf{MMLU-STEM} \\ 
    & Acc. & Acc. \\
    \midrule
    Base Models \\ \midrule
    % Qwen2-30B-A3B & 81.5 \\
    % Qwen2.5-Plus & 85.6 \\
    % Qwen2.5-Turbo & 76.1 \\
    % Qwen2.5-0.5B Chat & 21.5 \\
    % Qwen2.5-0.5B & 20.3 \\
    % Qwen2.5-1.5B & 45.1 \\
    % Qwen2.5-3B & 56.3 \\
    Qwen2.5-7B & 59.9 & 69.6 \\
    DeepSeek-Distilled-Qwen-14B & 87.5& 90.7\\
    % Qwen2.5-14B & 78.2 \\
    % Qwen2.5-32B & 84.5 \\
    % Qwen2.5-72B & 86.3 \\
    % Qwen3-0.6B & 41.5 \\
    % Qwen3-1.7B & 54.5 \\
    Qwen3-4B & 89.0 & 91.1 \\
    Qwen3-8B & 90.9 & 92.4\\
    Qwen3-14B & \textbf{91.9} & \textbf{94.0} \\
    % Qwen3-235B-A22B & 88.9 \\
    % Qwen3-32B & 87.4 \\
    % \midrule
    % InternVL \\ \midrule
    % InternVL3-1B & 34.5 \\
    % InternVL3-2B & 52.0 \\
    % InternVL3-8B & 77.4 \\
    % InternVL3-14B & 82.5 \\
    % InternVl3-38B & 87.7 \\
    % InternVL3-78B & 85.2 \\
    % \midrule
    % Hermes \\ \midrule
    % Hermes-3-8B & 52.9 \\
    % Hermes-3-70B & 67.8 \\
    % Hermes-3-405B & 75.4 \\
    % \midrule
    % Skywork-13B & 41.7 \\
    % InternLM-20B & 62.1 \\
    % Baichuan-2-13B & 49.0 \\
    % Falcon-180B & 54.0 \\
    % Yi-6B & 42.8 \\
    % Yi-34B & 54.3 \\
    % Mistral-7B & 55.7 \\
    % DSV-3-Small & 41.4 \\ 
    % DSV2-Lite & 44.1 \\
    % Moonlight-A@1.2T & 45.3 \\
    % Moonlight@1.2T & 43.2 \\
    % Gemma-3-1B & 28.1 \\
    % Gemma-3-4B & 51.7 \\
    % Gemma-3-12B & 74.3 \\
    % Gemma-3-27B & 80.0 \\
    % Llama-3-8B & 57.7 \\
    % Llama-3.1-8B-Instruct & 48.8 \\
    % Llama-3.1-70B-Instruct & 69.2 \\
    % Llama-3.1-405B-Instruct & 76.3 \\
    % Llama 3.2-3B & 46.8 \\
    % Llama-4-Scout & 82.4 \\
    % Llama-4-Maverick & 83.6 \\
    Phi-4-R$^+$ & 90.8 & 92.9 \\
    \midrule
    Existing Reward Models \\ \midrule
    Prometheus-7B-v2.0 & 54.0 & 56.5 \\ 
    Selene-1-Mini-Llama-3.1-8B & 58.2 & 65.2 \\
    RISE-Judge-Qwen2.5-7B & 63.1 & 76.9 \\
    RISE-Judge-Qwen2.5-32B & \textbf{82.8} & \textbf{89.4} \\
    \midrule
    $\shortmethodname$ Models (Ours) \\ \midrule
    $\shortmethodname\textsc{-Qwen3-4B-LoRA-4K}$ & 89.0 & 92.1 \\
    $\shortmethodname\textsc{-Qwen3-4B-LoRA-14K}$ & 88.9 & 92.2 \\
    $\shortmethodname\textsc{-Qwen3-4B-4K}$ & 88.8 & 91.8 \\
    $\shortmethodname\textsc{-Qwen3-4B-14K}$ & 89.3 & 92.0\\
    $\shortmethodname\textsc{-Qwen3-8B-LoRA-4K}$ & 90.8 & 93.5\\
    $\shortmethodname\textsc{-Qwen3-8B-LoRA-14K}$ & 90.8 & 93.6 \\
    $\shortmethodname\textsc{-Qwen3-8B-4K}$ & 90.7 & 93.3\\
    $\shortmethodname\textsc{-Qwen3-8B-14K}$ & 90.7 & 93.6 \\
    $\shortmethodname\textsc{-Qwen3-14B-LoRA-4K}$ & 91.7 & \textbf{94.8}\\
    $\shortmethodname\textsc{-Qwen3-14B-LoRA-14K}$ & 91.9 & 94.5 \\
    $\shortmethodname\textsc{-Qwen3-14B-4K}$ & 92.1 & 94.6\\
    $\shortmethodname\textsc{-Qwen3-14B-14K}$ & 92.1 & \textbf{94.8}\\
    $\shortmethodname\textsc{-Phi-4-R$^+$-14B-LoRA-4K}$ & 91.4 & 93.3 \\
    $\shortmethodname\textsc{-Phi-4-R$^+$-14B-LoRA-14K}$ & 91.3 & 93.5 \\
    $\shortmethodname\textsc{-Phi-4-R$^+$-14B-4K}$ & 91.2 & 93.6 \\
    $\shortmethodname\textsc{-Phi-4-R$^+$-14B-14K}$ & \textbf{92.2} & 94.4\\
    $\shortmethodname\textsc{-Qwen2.5-7B-LoRA-4K}$ & 71.7 & 81.8 \\
    $\shortmethodname\textsc{-Qwen2.5-7B-4K}$ & 79.8 & 86.4 \\
    $\shortmethodname\textsc{-Qwen2.5-7B-14K}$ & 81.1 & 88.3 \\
    $\shortmethodname\textsc{-DeepSeek-Distilled-Qwen-14B-LoRA-4K}$ & 89.9 & 91.9 \\
    $\shortmethodname\textsc{-DeepSeek-Distilled-Qwen-14B-LoRA-14K}$ & 90.0 & 92.2 \\
    $\shortmethodname\textsc{-DeepSeek-Distilled-Qwen-14B-4K}$ & 91.3 & 92.9 \\
    $\shortmethodname\textsc{-DeepSeek-Distilled-Qwen-14B-14K}$ & 91.1 & 93.0 \\
 \midrule
    Propretiary Models \\ \midrule
    % GPT-3.5 & 70.1 \\
    % GPT-4 & 86.7 \\
    GPT-4.1 mini & 91.0 & 93.3 \\
    GPT-o4 mini & 93.2 & 95.3 \\
    GPT-5 mini & \textbf{95.0} & \textbf{96.5} \\
    % DeepSeek-Base-7B & 39.5 \\
    % DeepSeek-Chat-7B & 42.3 \\
    % DeepSeek-Base-67B & 68.7 \\
    % DeepSeek-Chat-67B & 71.7 \\
    % DeepSeek-Coder-Base-v1.5-7B & 55.5 \\
    % DeepSeekMath-Base-7B & 59.5 \\
    % DeepSeek-V2 & 78.8 \\
    % DeepSeek-V3 & 86.2 \\
    DeepSeek-R1 & 94.0 & 96.2 \\
    \bottomrule
    \end{tabular}
    % }
\vspace{-2mm}
\end{table*}

\begin{table*}[!ht]%[!ht]
\centering
\caption{Comparison of existing models with $\shortmethodname{}$ on XSUM and FeedbackBench. \textbf{Bolded numbers} indicate the best-performing results within each group section independently.}
\resizebox{\textwidth}{!}{
    \begin{tabular}{lcccc}
    \toprule
    \textbf{Models} & \multicolumn{3}{c}{\textbf{XSUM}} & \textbf{FeedbackBench}\\ 
    & Acc. &  \multicolumn{2}{c}{Kendall Tau} & Kendall Tau\\ 
    & Faithfulness & Coherence & Relevance & \\
    \midrule
    Base Models \\ \midrule
    Qwen2.5-7B & 57.3 & -0.04 & 0.30 & 0.66 \\
    DeepSeek-Distilled-Qwen-14B & 65.0 & 0.26 & 0.28 & 0.63 \\
    Qwen3-4B & \textbf{75.9} &0.17 & 0.25 & 0.67 \\
    Qwen3-8B& 72.5 &0.28 &0.33 & 0.67  \\
    Qwen3-14B & 70.2 &0.31 & \textbf{0.34} & 0.67  \\
    Phi-4-R$^+$-14B & 64.0 & \textbf{0.46} &0.30 & \textbf{0.70} \\
    % ArmoRM & ? & ? & ? & ? & ? & ? \\
    % Ada-Instruct-v1-350M & - & 3.90 & 3.87 \\
    % Ada-Instruct-v1-350M$^\diamond$ & - & 3.54 & 3.07 \\
    % Curie-Instruct-v1-6.7B & - & 4.27 & 4.34 \\
    % Curie-Instruct-v1-6.7B$^\diamond$ & - & 4.28 & 3.80 \\
    % Davinci-Instruct-v2-175B & - & 4.41 & 4.28 \\
    % Davinci-Instruct-v2-175B$^\diamond$ & - & 4.83 & 4.33 \\
    % Anthropic-LM-52B$^\diamond$ & - & 4.77 & 4.14 \\
    % Cohere-XL-52.4B$^\diamond$ & - & 4.79 & 4.00 \\
    % GLM-130B$^\diamond$ & - & 4.72 & 4.12 \\
    % OPT-175B$^\diamond$ & - & 4.80 & 4.01 \\
    \midrule
    Existing Reward Models \\ \midrule
    UniEval & \textbf{84.3} & 0.07 & 0.03 & - \\
    Prometheus-7B-v2.0 & 60.7 & 0.12 & 0.16 & \textbf{0.79}\\
    Selene-1-Mini-Llama-3.1-8B & 56.4 & 0.16 & 0.36 & 0.78\\
    RISE-Judge-Qwen2.5-7B & 66.4  & 0.29 & 0.32 & 0.68 \\
    RISE-Judge-Qwen2.5-32B & 71.0 & \textbf{0.30} & \textbf{0.39} & 0.74 \\ \midrule
    % Qwen3 4B & 76.3 & 0.17 & 0.26 & 0.66 & 85.0 & 87.2 \\
    % Qwen3 8B & 72.1 & 0.24 & 0.32 & 0.66 & 92.0 & 93.7 \\
    % Qwen3 14B & 70.9 & 0.34 & 0.42 & 0.66 & 92.5 & 94.8  \\ \midrule
    $\shortmethodname$ Models (Ours) \\ \midrule
    $\shortmethodname\textsc{-Qwen3-4B-LoRA-4K}$ & \textbf{70.8} & 0.12 & 0.26 & 0.63 \\
    $\shortmethodname\textsc{-Qwen3-4B-LoRA-14K}$ & 70.7 & 0.12 & 0.26 & 0.64 \\
    $\shortmethodname\textsc{-Qwen3-4B-4K}$ & 66.8 & 0.23 & 0.27 & 0.63 \\
    $\shortmethodname\textsc{-Qwen3-4B-14K}$ & 66.7 & 0.25 & 0.31 & 0.63 \\
    $\shortmethodname\textsc{-Qwen3-8B-LoRA-4K}$ & 67.7 & 0.22 & 0.32 & 0.65 \\
    $\shortmethodname\textsc{-Qwen3-8B-LoRA-14K}$ & 69.6 & 0.24 & 0.31 & 0.67 \\
    $\shortmethodname\textsc{-Qwen3-8B-4K}$ & 68.0 & 0.36 & 0.31 & 0.66 \\
    $\shortmethodname\textsc{-Qwen3-8B-14K}$ & 65.8 & 0.37 & 0.32 & \textbf{0.71} \\
    $\shortmethodname\textsc{-Qwen3-14B-LoRA-4K}$ &  67.8 & 0.26 & 0.35 & 0.64 \\
    $\shortmethodname\textsc{-Qwen3-14B-LoRA-14K}$ &  69.2 & 0.24 & 0.34 & 0.65\\
    $\shortmethodname\textsc{-Qwen3-14B-4K}$ & 67.8 & 0.34 & 0.34 & 0.68 \\
    $\shortmethodname\textsc{-Qwen3-14B-14K}$ & 68.5 & 0.33 & \textbf{0.36} & \textbf{0.71} \\
    $\shortmethodname\textsc{-Phi-4-R$^+$-14B-LoRA-4K}$ & 64.8 & \textbf{0.45} & 0.31 & 0.69 \\
    $\shortmethodname\textsc{-Phi-4-R$^+$-14B-LoRA-14K}$ & 61.8 & 0.40 & 0.30 & 0.68 \\
    $\shortmethodname\textsc{-Phi-4-R$^+$-14B-4K}$ & 67.5 & 0.36 & 0.30 & 0.69 \\
    $\shortmethodname\textsc{-Phi-4-R$^+$-14B-14K}$ & 67.3 & 0.35 & 0.34 & 0.67 \\
    $\shortmethodname\textsc{-Qwen2.5-7B-LoRA-4K}$ & 52.8 & 0.14 & 0.21 & 0.48 \\
    % $\shortmethodname\textsc{-Qwen2.5-7B-LoRA-14K}$ &  \\
    $\shortmethodname\textsc{-Qwen2.5-7B-4K}$ & 65.1 & 0.29 & 0.29 & 0.64 \\
    $\shortmethodname\textsc{-Qwen2.5-7B-14K}$ & 67.5 & 0.33 & 0.34 & 0.69 \\
    $\shortmethodname\textsc{-DeepSeek-Distilled-Qwen-14B-LoRA-4K}$ & 58.4 & 0.21 & 0.31 & 0.64 \\
    $\shortmethodname\textsc{-DeepSeek-Distilled-Qwen-14B-LoRA-14K}$ & 59.9 & 0.37 & 0.32 & 0.66 \\
    $\shortmethodname\textsc{-DeepSeek-Distilled-Qwen-14B-4K}$ & 61.9 & 0.39 & 0.31 & 0.69 \\
    $\shortmethodname\textsc{-DeepSeek-Distilled-Qwen-14B-14K}$ & 64.3 & 0.40 & 0.34 & \textbf{0.71} \\
    \midrule
    Proprietary Models \\ \midrule
    % GPT-3-350M & - & 2.03 & 1.90 \\
    % GPT-3-6.7B & - & 3.16 & 3.39 \\
    % GPT-3-6.7B$^\diamond$ & - & 4.19 & 3.36 \\
    % GPT-3-175B & - & 2.78 & 3.52 \\
    % GPT-3-175B$^\diamond$ & - & 4.69 & 4.03 \\
    GPT-4.1 mini & \textbf{72.6} & 0.07 & 0.38 & 0.69\\
    GPT-o4 mini & 69.1 & 0.16 & 0.30 & 0.66 \\
    GPT-5 mini & 68.7 & \textbf{0.42} & \textbf{0.39} & 0.62 \\
    DeepSeek-R1 & 60.4 & 0.35 & 0.38 & \textbf{0.72}\\
    \bottomrule
    \end{tabular}
    }
\label{tab:XSUM-supplementary}
\vspace{-2mm}
\end{table*}

\subsection{On-Policy Optimization Experiments}
\label{sec:on-policy-opt-exps}

In addition to the off-policy optimization result in Table~\ref{results-rlhf}, Table~\ref{tab:results-grpo-ablation} presents the performance of different on-policy optimization methods using the $\shortmethodname$ reward models on MT-Bench and WildBench. We observe that GRPO's performance scales favorably with sample size $n$ for the 8B GRM, with $n = 16$ yielding the best results across nearly all WildBench categories as well as the best MT-Bench score. Notably, REINFORCE++ underperforms across both GRMs relative to GRPO, and in some cases barely improves over the initial policy, suggesting that variance reduction from group-relative sampling in GRPO is particularly beneficial when rewards come from a generative model rather than a scalar RM. However, the 14B-LoRA GRM does not exhibit the same clean scaling behavior with $n = 8$ on most WildBench dimensions before degrading at $n = 16$, and MT-Bench scores fluctuate non-monotonically across all settings for this GRM.

Comparing against DPO-based preference optimization results, we find that DPO with comparably sized reward models generally achieves stronger and more stable performance than our on-policy GRPO variants. We attribute this to reward calibration challenges inherent to using a generative reward model under GRPO; unlike DPO, which sidesteps explicit advantage estimation through a pairwise likelihood objective, GRPO requires well-calibrated scalar rewards to compute meaningful advantages, and miscalibration in the generative RM's scoring likely introduces noisy or biased advantage estimates that limit and destabilize policy improvement, particularly at larger $n$ and with the LoRA-adapted 14B GRM. In addition, we provide the compute report at Table~\ref{tab:grpo-wall-time}, showcasing that DPO is also much easier to optimize in faster iteration. We consider developing better reward calibration strategies for generative RMs under GRPO as promising future work.

\begin{table*}[!t]
\centering
\caption{On-Policy Optimization results comparing REINFORCE++ and GRPO (with varying sample size $n$) across two Generative Reward Models (GRMs), $\shortmethodname$\textsc{-Qwen3-8B-14k} and $\shortmethodname$\textsc{-Qwen3-14B-LoRA-4k}. \textbf{Bolded numbers} indicate the best performing results, while \underline{underlined numbers} indicate the second-best performing results.}
\resizebox{\textwidth}{!}{
    \begin{tabular}{lc|ccccccc}
    \toprule
    Method & MT-Bench & & & WildBench \\
    & Overall & Overall & Creative Plan. & Data Analy. & Info. Seek. & Coding & Math \\ \midrule
    Llama 3.2 3B Instruct (Init Policy) & 5.75 & 12.66 & 36.07 & 18.74 & 1.51 & 27.28 & -2.64 \\
    \midrule
    GRM: $\shortmethodname$\textsc{-Qwen3-8B-14k} \\
    \midrule
    + REINFORCE++ & 5.85 & 12.62 & 37.16 & 17.79 & 2.14 & 26.73 & -2.45 \\
    + GRPO ($n=4$) & 5.66 & 18.72 & 41.81 & 24.84 & 6.27 & 36.04 & 2.92 \\
    + GRPO ($n=8$) & 5.82 & \underline{21.19} & \underline{42.38} & \underline{26.85} & \underline{8.73} & \underline{37.48} & \underline{7.26} \\
    + GRPO ($n=16$) & \textbf{6.07} & \textbf{22.68} & \textbf{45.43} & \textbf{28.88} & \textbf{9.84} & \textbf{39.06} & \textbf{7.83} \\
    \midrule
    GRM: $\shortmethodname$\textsc{-Qwen3-14B-LoRA-4k} \\
    \midrule
    + REINFORCE++ & 5.81 & 12.96 & 37.26 & 18.48 & 2.46 & 28.02 & -2.92 \\
    + GRPO ($n=4$) & 5.83 & 15.87 & 40.67 & 21.49 & 5.56 & 31.09 & -0.57 \\
    + GRPO ($n=8$) & 5.78 & 19.32 & 41.81 & 25.65 & 8.65 & 33.47 & 4.06 \\
    + GRPO ($n=16$) & \underline{5.87} & 18.65 & 42.69 & 24.99 & 6.35 & 35.15 & 2.64 \\
    \bottomrule
    \end{tabular}
}
\label{tab:results-grpo-ablation}
\end{table*}

\begin{table*}[!th]
\centering
\caption{Wall-clock time of different optimization methods using 4$\times$H100 GPUs.}
\begin{tabular}{lc}
\toprule
\textbf{Method} & \textbf{Wall Time (4$\times$H100)} \\
\midrule
DPO          & $\sim$4.06 hrs \\
REINFORCE++  & $\sim$4.59 hrs \\
GRPO ($n=4$) & $\sim$16.03 hrs \\
GRPO ($n=8$) & $\sim$35.39 hrs \\
GRPO ($n=16$)& $\sim$87.95 hrs \\
\bottomrule
\end{tabular}
\label{tab:grpo-wall-time}
\end{table*}

\subsection{SFT vs. RLVR: Performance Comparisons and Cost Trade-offs}
\label{sec:app-sft-vs-rl}

We include additional comparisons to better contextualize its benefits relative to a strong SFT baseline. All experiments are run with LoRA. We compare SFT and RLVR under two data settings: 14k (full dataset) and a 7k-7k split where the data budget is divided equally between SFT and RLVR.

Overall, Table~\ref{tab:sft-rlvr-results} shows that RLVR provides modest but consistent gains over SFT under the full 14k setting across RM-Bench (83.8 vs. 83.5), BBH (92.3 vs. 91.9), MMLU-STEM (94.9 vs. 94.5), XSUM (70.7 vs. 69.2), and FeedbackBench (faithfulness, coherence, relevance). These results confirm that RLVR can further refine an already strong SFT model, though the margins are incremental rather than transformative. Importantly, the SFT 14k model itself already achieves highly competitive performance across all benchmarks, closely matching RLVR 14k in most cases, demonstrating that a well-constructed SFT pipeline captures most of the achievable performance in this setting. This is further supported by the 7k-7k split, where SFT 7k $\rightarrow$ RLVR 7k performs on par with SFT 14k alone across most benchmarks, suggesting that splitting the data budget between SFT and RLVR offers no clear advantage over simply training SFT on the full dataset.

Finally, these marginal gains from RLVR come at substantial computational overhead as shown in Table~\ref{tab:sft-rlvr-hrs-comparison}, in which RLVR 14k requires $\sim$112 GPU hours compared to $\sim$8 hours for SFT 14k. Overall, these findings suggest that SFT is not only a strong baseline but also a highly cost-effective approach, with RLVR offering additional but diminishing returns at significantly higher computational expense.

\begin{table*}[!th]
\caption{Benchmark results for different model variants.}
\centering
\resizebox{\textwidth}{!}{%
\begin{tabular}{lcccccccc}
\toprule
 & \textbf{RewardBench} & \textbf{RM-Bench} & \textbf{BBH} & \textbf{MMLU-STEM} & \multicolumn{3}{c}{\textbf{XSUM}} & \textbf{FeedbackBench} \\
\cmidrule(lr){6-8}
 & & & & & Faith. & Coh. & Rel. & \\
\midrule
SFT 14k                      & 90.2 & 83.5 & 91.9 & 94.5 & 69.2 & 0.24 & 0.34 & 0.65 \\
RLVR 14k                     & 89.7 & 83.8 & 92.3 & 94.9 & 70.7 & 0.33 & 0.35 & 0.67 \\
SFT 7k $\rightarrow$ RLVR 7k & 87.0 & 75.7 & 87.1 & 92.1 & 67.6 & 0.22 & 0.17 & 0.41 \\
SFT 7k                       & 87.0 & 75.6 & 87.3 & 92.1 & 67.3 & 0.21 & 0.17 & 0.39 \\
\bottomrule
\end{tabular}
}
\label{tab:sft-rlvr-results}
\end{table*}

\begin{table*}[!th]
\caption{GPU hours between SFT and RLVR settings.}
\centering
\begin{tabular}{lr}
\toprule
 & \textbf{H100 GPU hours}\\
\midrule
SFT 14k                      & $\sim$8 hrs \\
RLVR 14k                     & $\sim$112 hrs \\
SFT 7k $\rightarrow$ RLVR 7k & $\sim$40 hrs \\
SFT 7k                       & $\sim$4 hrs \\
\bottomrule
\end{tabular}
\label{tab:sft-rlvr-hrs-comparison}
\end{table*}

\end{document}